\documentclass[final]{siamltex}

\usepackage{latexsym,amssymb,lastpage}
\usepackage{graphicx,amsfonts}
\usepackage{times,mathptmx,bm,amsmath}
\usepackage{dcolumn}
\usepackage{epsfig}

\usepackage{tipa}

\title{Exemplar Dynamics and Sound Merger in Language
\thanks{The author was supported by a
Discovery Grant awarded by the Natural Sciences and Engineering
Research Council of Canada and a Tier II Canada
Research Chair. A significantly abbreviated version of this work appeared in the Proceedings of the 36th Annual Conference of the Cognitive Science Society 2014.}}

\author{P. F. Tupper\thanks{Department of Mathematics, Simon Fraser University, Burnaby BC, V5A 1S6, Canada ({\tt pft3@sfu.ca}).} }

\begin{document}

\maketitle

\begin{abstract}
We develop a model 
of \emph{phonological contrast} in natural language. Specifically, our model describes the maintenance of contrast between different words in a language, and the elimination of such contrast when sounds in the words merge.  An example of such a contrast is that provided by the two vowel sounds \emph{i} and \emph{e}, which distinguish pairs of words such as \emph{pin} and \emph{pen} in most dialects of English. 
We model language users' knowledge of the pronunciation of a word as consisting of collections of labeled exemplars stored in memory. 
  Each exemplar is a detailed memory of a particular utterance of the word in question.  In our model an exemplar is represented by one or two phonetic variables along with a weight indicating how strong the memory of the utterance is.   Starting from an exemplar-level model we derive integro-differential equations for the evolution of exemplar density fields in phonetic space. Using these latter equations we investigate under what conditions two sounds merge, thus eliminating the contrast. Our main conclusion is
that for the preservation of phonological contrast, it
is necessary that anomalous utterances of a given word
are discarded, and not merely stored in memory as an exemplar
of another word.
\end{abstract}



\pagestyle{myheadings}
\thispagestyle{plain}
\markboth{P. F. TUPPER}{EXEMPLAR DYNAMICS AND SOUND MERGER IN LANGUAGE}

\section{Introduction}


Phonology is the study of  how sequences of meaningless sounds are coordinated in natural languages to make meaningful speech \cite{gussenhoven,silverman}.
A key idea in phonology is that of \emph{phonological contrast}. It is the contrast between the sound of different portions of speech that allows meaning to be conveyed.  For example, if I am asked ``What will you have to drink?" I can respond by saying ``coffee" or ``tea", sequences of sounds that have no intrinsic meaning but allow me to convey meaning to you because they form words.   There are many perceptible contrasts between these two words, and this allows information to be conveyed to the hearer. Phonology studies how such contrasts are organized. 
For example, an  important concept in phonology is that of the {\it minimal contrastive pair}. This is a pair of words that differ in only one sound, such as the pairs ``bit, bet" or  ``can, ban".


One observation about contrast in languages is that sometimes words that contrasted in the past do not do so any more \cite{hock_joseph}. This is known as a \emph{sound merger}. Sound mergers are most noticeable when the merger has occurred in one dialect of a language but not another.
  An example is the well-known {\sc cot--caught} merger. Historically, the words ``cot" and ``caught" formed a minimally contrastive pair, having vowels that were pronounced distinctly by English speakers. However, in many dialects of English today (including the dialect of the author) these words are indistinguishable: the vowel sounds in ``cot" and ``caught" have merged.  Another example is the {\sc near--square} merger in New Zealand English, in which the words ``beer" and ``bear" are not distinguishable from their pronunciation \cite{hickey}.

How are these contrasts and mergers quantified and studied? The acoustic signal of a spoken utterance is extremely complicated, and at a fine detailed level no two utterances ever spoken are identical. However, through a variety of methods linguists have learned to identify the features of the speech signal that are the most salient to hearers and that do most to establish contrast between different words.  
One well-studied  family of contrasts is  provided by  vowel sounds.
Linguists have learned that the most important features of vowels are two quantities known as F1 and F2, which we now explain \cite{ladefoged}.

 To a rough approximation, the sound of a vowel is a periodic signal with a base frequency  known in linguistics as F0 (the fundamental frequency).  F0 varies a great deal from speaker to speaker and its value is  not itself used to provide contrast between words in English. 
 In addition to the  base frequency F0 there are harmonics, consisting of integer multiples of F0. Within the harmonics,
 the frequencies at which there is a peak in  amplitude  are known as \emph{formants}. The frequency of the first formant is known as F1, the frequency of the second formant as F2, and so on. 
  The first two formants, F1 and F2, are usually the most important cues used to distinguish vowels in spoken languages \cite{ladefoged}.

\begin{figure}
\begin{center}
\includegraphics[width=12cm]{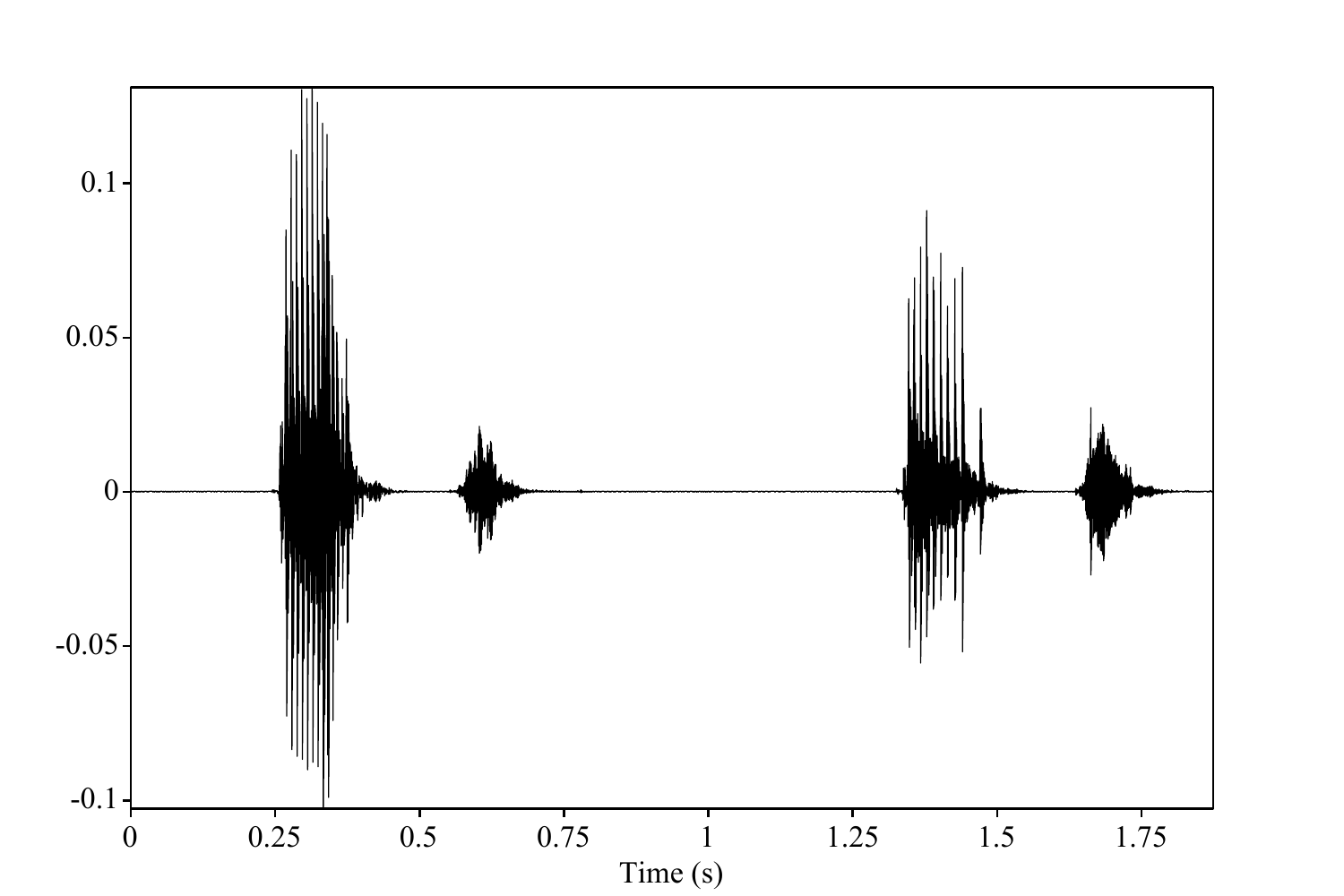}
\includegraphics[width=12cm]{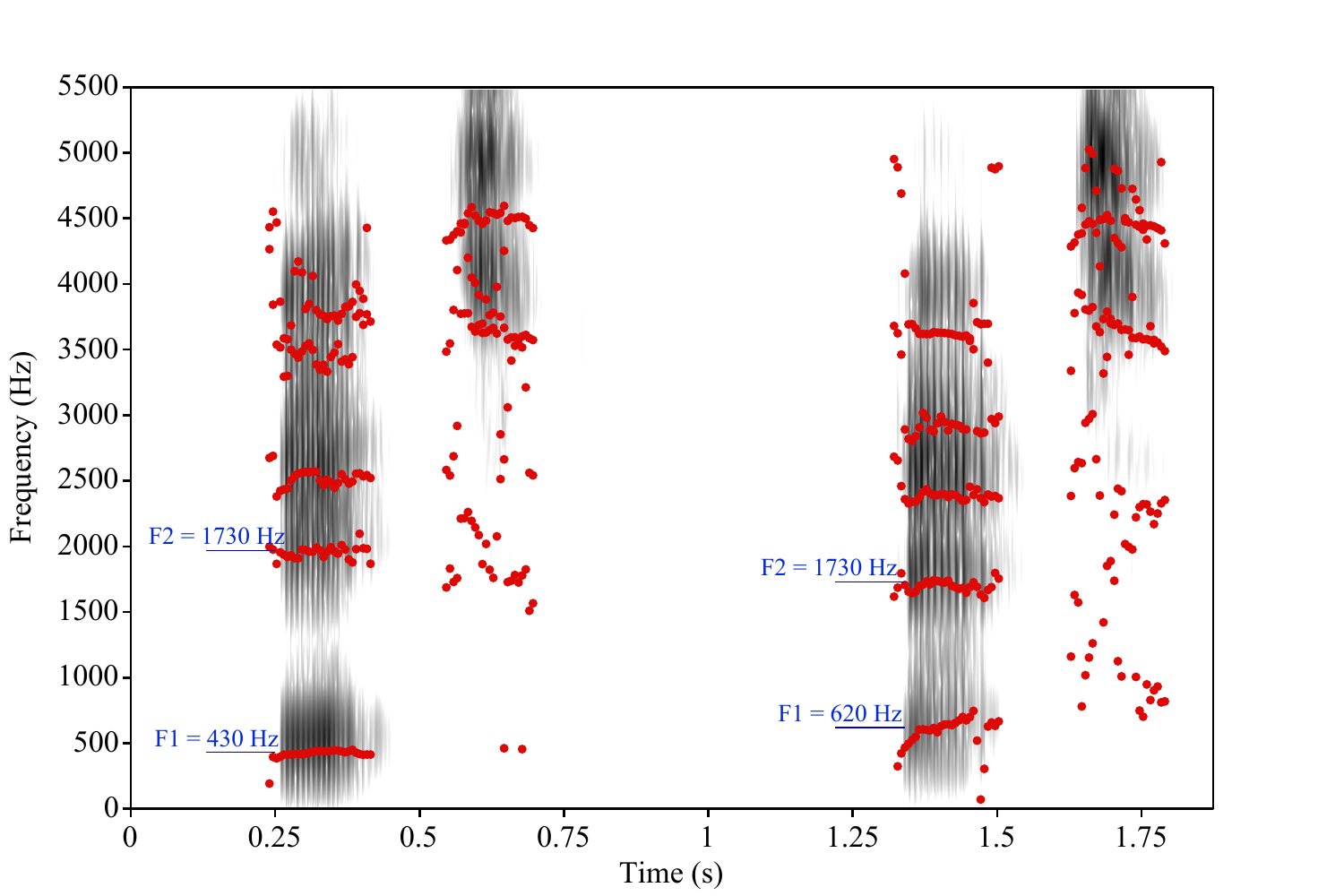}
\end{center}
\caption{  \label{fig:bit_bet}  The author saying ``bit bet". Data plotted using the software Praat \cite{praat}. The top shows the sound signal in a raw form as amplitude vs time. The bottom shows a spectrogram, as frequency versus time. The darkness of the plot shows the intensity of the sound at that frequency at that point in time.
}
\end{figure}

We demonstrate these points with some data. In Figure~\ref{fig:bit_bet} we show plots of the author saying the words ``bit, bet", generated with the help of the phonetics software package Praat \cite{praat}.  The top plot shows the raw acoustic signal. The first pair of visible disturbances corresponds to the word ``bit". The first disturbance is mostly the vowel sound ``i" and the second smaller disturbance is the release of the ``t" sound. The period of quiet between the two bumps is when no air is escaping the mouth while making the ``t". The second pair of bumps is similarly the word ``bet".  In order to see the contrast between the words, in the second plot we show the spectrogram of the sound signals. The vertical axis gives frequency. Darkness in the plot corresponds to greater amplitude at a given time and frequency. During the duration of the  vowel sounds Praat has computed the frequency of the formants as a function of time and plotted them in red. Taking the average of the formant frequency over the duration of the vowel, we obtain what we identify as the formant frequency for the vowel, indicated with a blue line and blue text in the plot. For the word ``bit" we obtain F1$= 430$ Hz, F2$=1970$ Hz. For the word ``bet" we obtain 
F1$= 620$ Hz, F2$=1730$ Hz. It is these differences in F1 and F2  that are primarily responsible for the words ``bit" and ``bet" sounding different.

\begin{figure}
\begin{center}
\includegraphics[width=12cm]{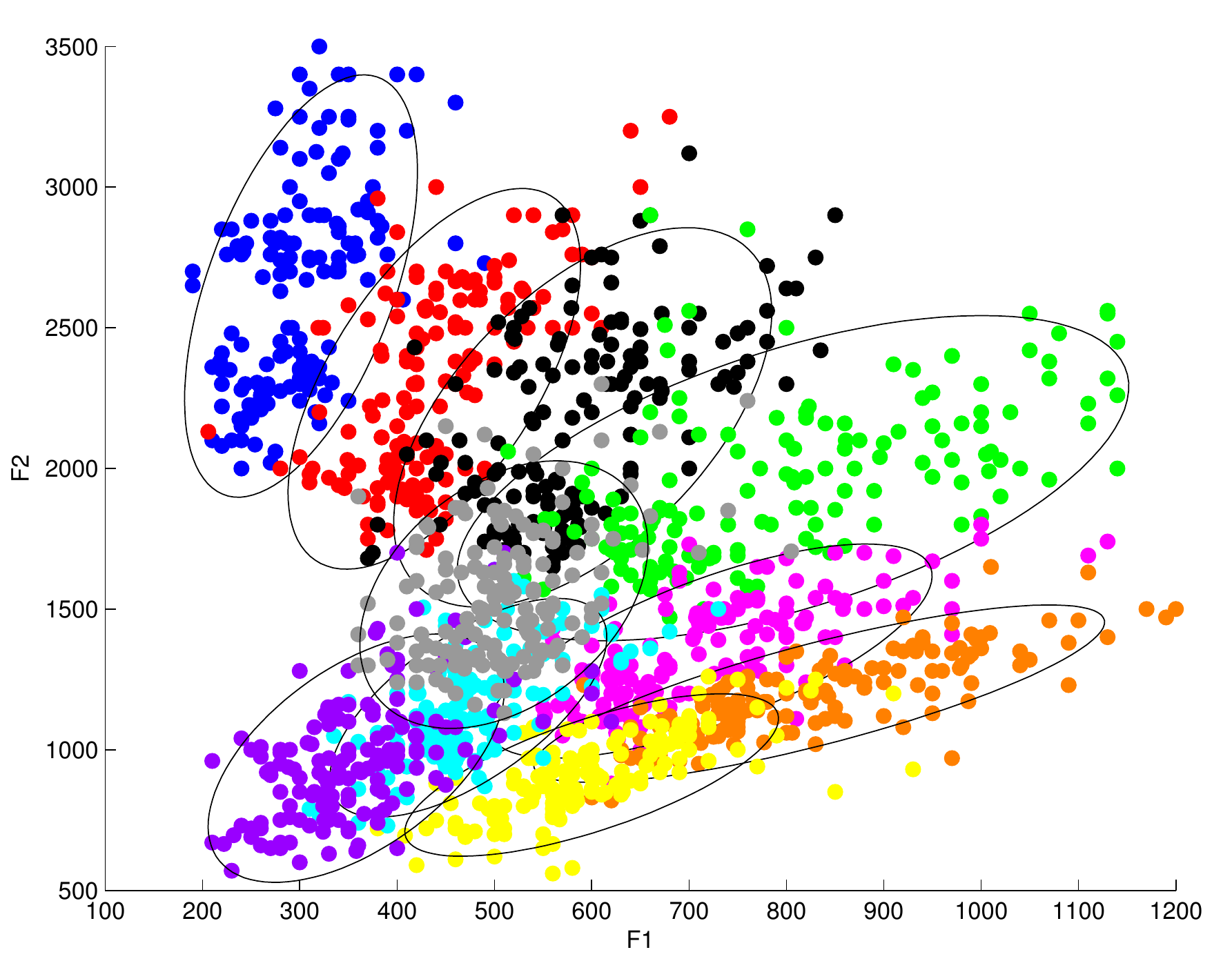}
\end{center}
\caption{  \label{fig:peterson_barney} 
Vowel data from the classic paper  \cite{peterson_barney}. Each point represents the F1 and F2 values of a recorded vowel. Data consisted of 1520 tokens from 76 speakers, including women and men, children and adults. The vowels are \textipa{i} (blue), \textipa{I} (red), \textipa{E} (black), \textipa{\ae} (green), \textipa{3} (grey), \textipa{2} (magenta), \textipa{A} (orange), \textipa{u} (purple), \textipa{U} (cyan), \textipa{O} (yellow).
 \cite{pierrehumbert_probability}}
\end{figure}

To a very rough approximation, all speakers in a homogeneous linguistic community use F1 and F2 the same way in order to distinguish vowels. Within speakers, there is variation in the way vowels are produced from utterance to utterance, but all instances of a  vowel occupy a similar region in F1-F2 space. To illustrate this Figure~\ref{fig:peterson_barney} shows the F1 and F2 of many vowels collected from 76 speakers of American English in 1952 \cite{peterson_barney}. 
Note that the vowels in this plot overlap in places. In reality, these vowels are likely disambiguated by other phonetic variables (such as F3) which are not shown in this plot \cite{pierrehumbert_probability}.

We have just described how the vowel system in languages such as English work. We mention two other major features of this system. First, over long periods of time the system is very stable in that the number of sounds and their location in phonetic space do not change much. When English speakers of today hear Alexander Graham Bell's recording of his voice from 1885, it is quite intelligible \cite{gray2013}. 
 This might be surprising given the imperfect and noisy way speech is transmitted from person to person and generation to generation of speakers. The location of vowels may shift within F1-F2 space (as we may say by noting the difference in pronunciation between British and American speakers \cite{ladefoged}), but for the most part the vowel system as a whole remains intact.
 Second, despite this stability, there occasionally are changes to the system, either through splits, where what was once a single sound becomes two, or through mergers, the focus of our discussion here.
 We need a model of the system that can account for its stability in most situations over long periods of time but also can explain when breakdowns in the system occur-- as they occasionally do-- in the form of mergers. 


%
%

One explanation for why the vowels in ``cot" and ``caught"
merged in some variants of English is that when the two words are spoken in the context
of a sentence it is easy to tell which is intended, since
``cot" is a noun and ``caught" is a verb. The merger is unlikely to result in any confusion. 
In contrast, a merger between
the pronunciations of the words ``can" and ``can't" could
make it difficult to communicate. 
Indeed, recent studies have shown that sound mergers occur more often in situations where the loss of the contrast provided by the two sounds is less harmful to communication \cite{bouchard-cote,wedel_kaplan_jackson}. What remains is to demonstrate the actual mechanism behind this phenomenon.
 In this paper we provide a detailed and explicit model of
this important phenomenon that does not rely on language users' higher-level goals (such as a desire to be intelligible), but just on how hearers classify what they hear as words. We demonstrate a simple model of how the pronunciation of words is stored in the mind that is
 sufficient to
explain why two sounds may merge in some linguistic contexts but not in others.


Our starting point is \emph{exemplar models}, which were first introduced in a non-linguistic context in the work of Nosofsky \cite{nosofsky1,nosofsky2}.  In studying how people performed categorization tasks, Nosofsky  hypothesized that each time a subject was exposed to a stimulus taken from some category, the subject stored a detailed memory of this single stimulus in their mind, which was called an \emph{exemplar}.  Exemplar theory became influential in linguistics through the work of Johnson \cite{johnson} and Pierrehumbert \cite{pierrehumbert_exemplar}. In particular, the latter paper \cite{pierrehumbert_exemplar} introduced the idea of 
\emph{exemplar dynamics}, in which the long term behaviours of word pronunciations within a population were modelled through the creation, interaction, and eventual fading away of exemplars in speakers' minds. Further work on exemplars dynamics in linguistics can be found in  \cite{wedel2006,ettlinger,blevins_wedel,wedel2012}.

\begin{figure}
\begin{center}
\includegraphics[width=12cm]{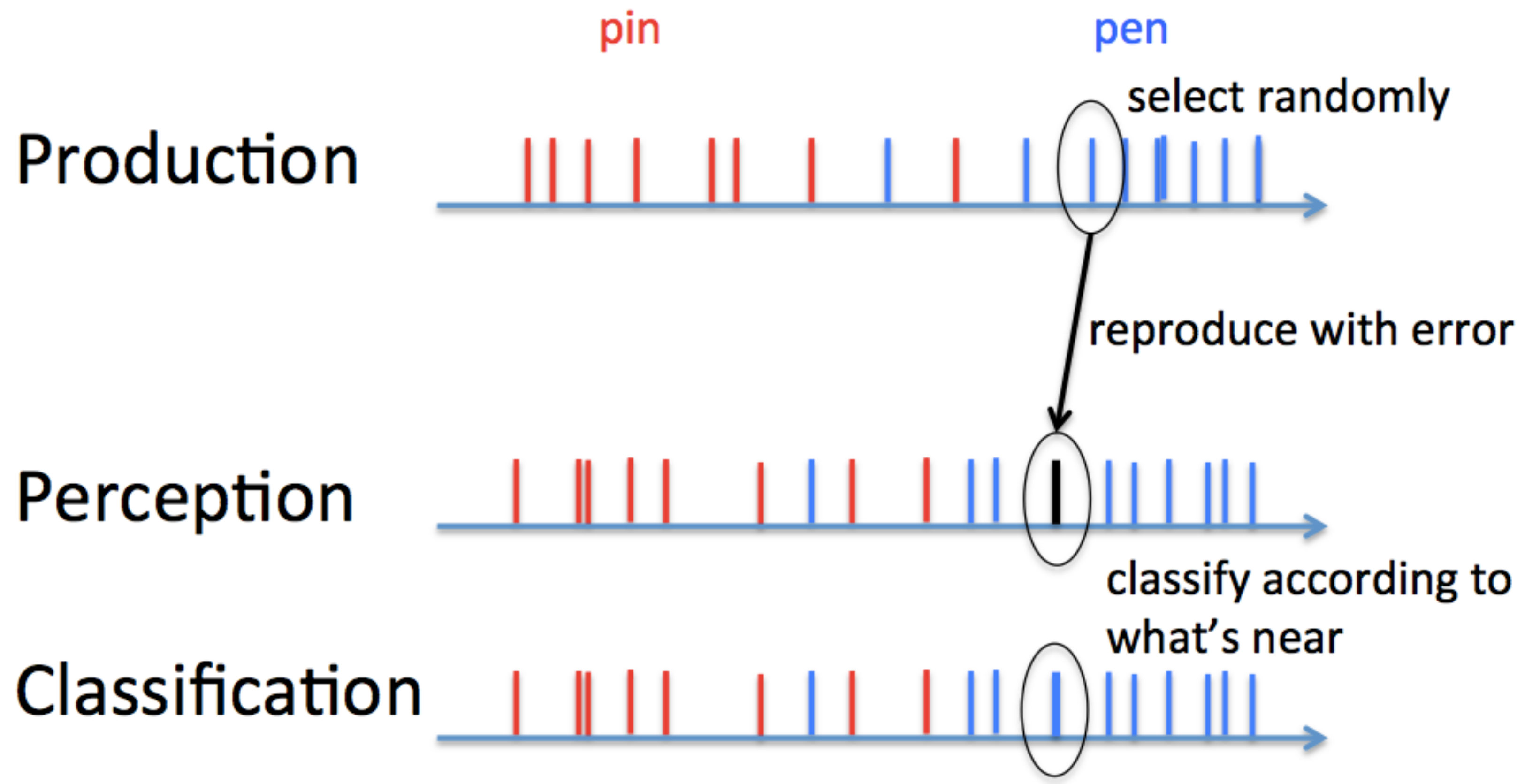}
\end{center}
\caption{  \label{fig:pangolin} Production by the speaker and reception by the hearer of the word ``pen". Though ``pin" and ``pen" are adjacent to each other phonetically, the word is correctly perceived by the hearer in this case. Top: the exemplars of the speaker, and the selected exemplar of the word ``pen". Middle: the exemplars of the hearer, and how the (corrupted) exemplar from the speaker is stored. Bottom: The exemplar is correctly classified as an exemplar of the word ``pen".}
\end{figure}

To give an idea of the basic structure of exemplar models in linguistics, in Figure~\ref{fig:pangolin} we show how such a model might handle the contrast between the words ``pin" and ``pen". We demonstrate a speaker saying the word ``pen" and that word being successfully communicated to the hearer.  In the top of the figure we show the speaker's representation of the two words ``pin" and  ``pen". In reality, the pronunciation of each word is specified by many phonetic variables; here we just show a single phonetic variable on the $x$-axis that
represents some aspect of the vowels  that  provide contrast between the two words.
 Exemplars of the word ``pin" are shown in red and exemplars of the word ``pen" are shown in blue.
The speaker decides to say the word ``pen" and so selects one of the (blue) exemplars of ``pen" uniformly at random (shown circled). This exemplar is produced by the speaker as an uttered word, but it is not a perfect duplication of the exemplar: noise and biases of various sorts perturb its phonetic value.  The  word is heard by the hearer and its phonetic value is compared to the hearer's other exemplars. This is shown in the middle, where the received exemplar is coloured black, as it has not been identified as being one word or another yet. Finally, based on the fact that the received exemplar is close to many of the (blue) exemplars of ``pen", and not close to many of the (red) exemplars of ``pin", the hearer classifies the word as an example of the word ``pen", as shown in the bottom of Figure~\ref{fig:pangolin}. 
 The model may include the speaker themselves storing the new exemplar in their memory or not.
This step completes one cycle of reproduction of an exemplar. 
   This cycle may then continue with the hearer in turn becoming the speaker, and the recently stored exemplar affecting their subsequent production. 
 
In this paper we describe  a particular exemplar dynamics model that elaborates on that shown in Figure~\ref{fig:pangolin}, and is built directly on the models of  \cite{pierrehumbert_exemplar} and \cite{blevins_wedel}. 
In our model every language-user has a small number of  words stored in their memory. We assume these words are preexisting, and the number of them is fixed.  Each word has associated with it a collection of exemplars that are memories of instances of that particular word being used. The exemplars of a word hold all the information the language-user has about the pronunciation of the word.
 Each exemplar is described by a scalar or vector $y$ describing one or more  phonetic variables (for example, in the case of vowels, F1 and F2), as well as a scalar weight $w$ that indicates how strong the memory of the exemplar is. All exemplars are initiated with a given weight $W_0$ when they are created  and then the weight decays with time.
New exemplars are created when a speaker selects one of their words to speak, selects an exemplar from that word's exemplar collection (randomly with probability proportional to  the exemplar weight), and utters it.
The exemplar is then received by the hearer who classifies it as  one of their words and stores the new exemplar in that word's exemplar collection.

The new exemplar is not an exact duplicate of the original exemplar. Either in the production of the exemplar by the speaker or in the reception and storage of the exemplar by the hearer (it doesn't matter which for our purposes) there are errors made.
The new exemplar's position is close to that of the original exemplar but undergoes some perturbations, which we now describe.

Firstly, we introduce a bias of the new phonetic values towards less extreme values. A similar effect is used  in the model  \cite{pierrehumbert_exemplar} and is denoted \emph{lenition}. In that paper, the phonetic value of an exemplar is decreased by a fixed amount every time it is reproduced. Psychologically, this corresponds to the undershooting of a  target \cite{lindblom1996}. For the long-time simulations we are interested in here, this choice would result in the phonetic values of all exemplars going to $- \infty$. Instead, following \cite{blevins_wedel,wedel2012} we introduce a bias towards the centre of perceptual space, indicating that speakers have a preference for producing exemplars that have neither too large nor too small phonetic values. We model this by postulating a preferred value of the phonetic variable $y$ which we indicate by $y^*$.  New exemplars have their phonetic value shifted  towards $y^*$ in proportion to their distance from it. We assume that $y_*$ is the same for all speakers and does not depend on the category the exemplar is copied from. Naturally this is a gross oversimplification as can be seen from the data in \cite{vallabha_tuller} where such biases are studied experimentally.

Secondly there is what Pierrehumbert \cite{pierrehumbert_exemplar} refers to as \emph{entrenchment}:
new exemplars produced from a given category are produced with their phonetic values biased towards the mean phonetic value of the exemplars already in this category. Unlike lenition, this bias depends on the other exemplars in the system, and so varies with the individual language user and the category in question.
As described below, we use a version of entrenchment that is simpler to work with, but has the same effect as those used in \cite{pierrehumbert_exemplar} and \cite{blevins_wedel}.

Finally, new exemplars are produced with noise, which we model as a mean-zero Gaussian random variable added to the phonetic variable of the exemplar after lenition and entrenchment have been added. 
 
As we describe in more detail in Section~\ref{sec:onecategory}, these factors of lenition, entrenchment, and noise are incorporated into a model of how the distribution of exemplars in a population of language-users changes over time. This in turn leads to predictions about how and under what conditions the pronunciations of words in the language change over time.

%

Despite their many successes, exemplar models as described here have some features that make them difficult to work with computationally. On the one hand, the computational expense of simulating an exemplar model grows with the number of exemplars in the system. Phenomena that may only emerge with millions or billions of exemplars may be missed in simulations of smaller systems. On the other hand, exemplar models involve randomness, through both the selection of exemplars to be reproduced, and in the errors in the production of new exemplars. This means that the result of simulations of exemplar models are random. In order to determine if a phenomenon in an exemplar simulation is a rare aberration or completely typical, it is necessary to do several exemplar simulations with the same parameters, adding to the cost of studying the system computationally.  

To overcome these difficulties we study our exemplar models in the limit where the system contains infinitely many exemplars. Instead of keeping track of variables for every exemplar in the system,  we use a continuous field variable describing the density of exemplars in phonetic space. This is quite analogous to the common procedure in mathematics and physics of deriving partial differential equations for particle density fields from laws for microscopic particle dynamics.
 In this limit, which we call the {\it field model}, rather than the location and weight of every exemplar in phonetic space, we consider fields which describe the density of exemplars of different types within phonetic space. We derive integro-differential equations for these exemplar density fields, which we refer to as field models for exemplar dynamics.   We study the field models in order to infer properties of the exemplar model in the limit of large numbers of exemplars.

Through simulations of exemplar models and the limiting field models, we show in this paper how the procedure by which a language-user classifies heard exemplars  determines whether or not the pronunciation of two words merges. We will see that if the language user always classifies heard exemplars accurately, no matter how they are produced, then the sounds in the words merge. 
 On the other hand, if there is the potential for exemplars to be misclassified, then the sounds will remain distinct, in agreement with the simulations of \cite{blevins_wedel}.  However, in contrast to \cite{blevins_wedel}, we also find that an important aspect of our model for predicting non-merger is that misclassified exemplars are rejected as unacceptable and not stored in memory as the wrong word.  This is the mechanism for prevention of sound merger proposed
by Labov in \cite[pp.\ 586--588]{labov}; see also \cite[Ch. 5]{silverman}.

To the best of our knowledge this is the first mathematical analysis of exemplar dynamics. However,
our work is related to that of Shipman, Faria, and Strickland \cite{Shipman}. These authors also propose differential equation models for the change of vowel pronunciation within phonetic space. Motivated by a desire to model  vowel shifts, the authors propose a model in which  every individual in the population has an age and a particular  pronunciation of a given vowel. Partial differential equations are derived for how the number of people with a particular age and  pronunciation changes with time, taking into account learning and social trends. Whereas we assume in our model that  all individuals in the population have the same rich mental representation of the pronunciation of words, Shipman et al.\  assume that within an individual the pronunciation of each vowel  is represented by a single scalar, but this scalar is different between different members of the population. In this sense our goals are complementary to those of \cite{Shipman} in that we provide a mechanism for why two vowels do not merge, whereas they take non-merger as a given.  

In Section~\ref{sec:onecategory} we describe our exemplar model and its limiting field model in the case of a single word characterized by a single phonetic variable.
In Section~\ref{sec:multiple} we extend our exemplar model to more than one word,  describing our model of  interaction of pronunciations of words through the classification of produced exemplars by hearers. We derive the analogous field model for the multiple word case. We explore how different models of the way hearers classify exemplars leads to either merger or non-merger. In Section~\ref{sec:twodim} we show some numerical experiments of our exemplar model and its field limit in the case of two phonetic variables, allowing a more direct, though qualitative, comparison with the data in Figure~\ref{fig:peterson_barney}. In Section~\ref{sec:functional_load}  we discuss what predictions our model makes for sound merger in real languages and tie our work to the literature on functional load \cite{wedel_kaplan_jackson,wedel_kaplan_jackson_langandspeech}.  In Section~\ref{sec:conclusion} we summarize our results and discuss directions for further research.

In order to have tractable models of the phenomena of interest, in all that follows we have made a simplifying assumption. Rather than modelling each individual in a population of speakers with their own exemplars, we pool all the exemplars in the population. (This choice is analogous to random mating in the Wright-Fisher model in population genetics.) This will not make a significant difference to our results if each individual in the population has the same words with the same statistical characteristics of exemplars associated with their words.
With this assumption, it does not matter whether just hearers or both speakers and hearers store a copy of new exemplars in their lexicon, the effect just being to change some parameters.
So for the sake of simplicity we make the former modelling assumption: that new exemplars are only stored by the hearer.

\section{Exemplar Dynamics and Field Models for a Single Word} \label{sec:onecategory}

In this section we first describe our exemplar model in the case of a single word with exemplars possessing a  single phonetic variable. The single word case is linguistically unrealistic; however it lays down the groundwork for the multiple word case covered in the next section.  In Subsection~\ref{subsec:onecat_exemp} we present the exemplar  model together with some numerical simulations. In Subsection~\ref{subsec:onecat_nonloc} we present equations for the dynamics of a continuum field of exemplar density with non-local interactions, which we call the \emph{field model}. 

\subsection{Exemplar Model: Single Word} \label{subsec:onecat_exemp}

Here we describe the exemplar model for the case of a single word, where the phonetic value of the exemplars are represented by a single scalar. 
At any point in time the state of the system consists of the positions of $n$ exemplars $y_1, \ldots, y_n \in \mathbb{R}$, and their respective weights $w_1, \ldots, w_n \geq 0$.
Initially there are $n(0)$ exemplars with fixed positions and weights. The number of exemplars at time $t$, $n(t)$ is non-decreasing, and increases  through the creation of new exemplars.
 Once an exemplar is  created, its position does not change and it is never removed from the system, though its weight decreases.  New exemplars are introduced to the system through a Poisson process in time with constant rate $\nu$. The initial weight of each new exemplar is $W_0$. The weight of all exemplars decay with rate $\lambda$, so that exemplars that have been around for long enough cease to have significant effect on the system.     Thus if exemplar $i$ is created at time $t_i$ at location $y_i$ we have
 \[
 w_i(t_i)=W_0, \ \ \ \frac{dw_i}{dt}= -\lambda w_i,
 \]
 for all $t \geq t_i$. (In simulations we delete exemplars whose weight drops below a certain low threshold in order to improve computational efficiency.) 
 The phonetic value $y$ of new exemplars is determined through two steps. First, one of the existing exemplars is selected with probability proportional to its weight. We imagine that this exemplar is being spoken by a speaker.  Let $z$ be the position of this exemplar. Let  $\bar{y}$ be the weighted mean position of all the exemplars:
\[
\bar{y}= \frac{\sum_{i=1}^n w_i y_i  }{\sum_{i=1}^n w_i}.
\]
We also postulate a preferred value of the phonetic value $y^*$. We let $\eta$ indicate a standard Gaussian random variable independent of all variables in the system. $\alpha$, $\beta$, and $\sigma$ are parameters we explain below. The new exemplar is produced with phonetic variables $y$:
  \[
  y = z + \alpha (\bar{y} -z) + \beta (y^* -z) + \sigma \eta.
  \]
For $\alpha >0$ the new exemplar is produced with a bias towards $\bar{y}$, the size of $\alpha$ indicating the strength of the bias. This is our version of entrenchment, a simpler version of what is used in  \cite{pierrehumbert_exemplar} and \cite{blevins_wedel}, 
 which we find to be more analytically tractable. In common with this more realistic versions, the effect is that exemplar production is biased towards the mean value of the exemplars in a category.
 For $\beta >0$ the new exemplar is produced with a bias towards $y^*$ the size of $\beta$ likewise indicating the size of the bias. This latter bias may be due to factors in the speaker's production or the hearer's perception.
Finally, noise is modelled by the term $\sigma \eta$ where  $\sigma$ is a parameter representing the magnitude of the noise.

\begin{figure}
\begin{center}
\includegraphics[width=12cm]{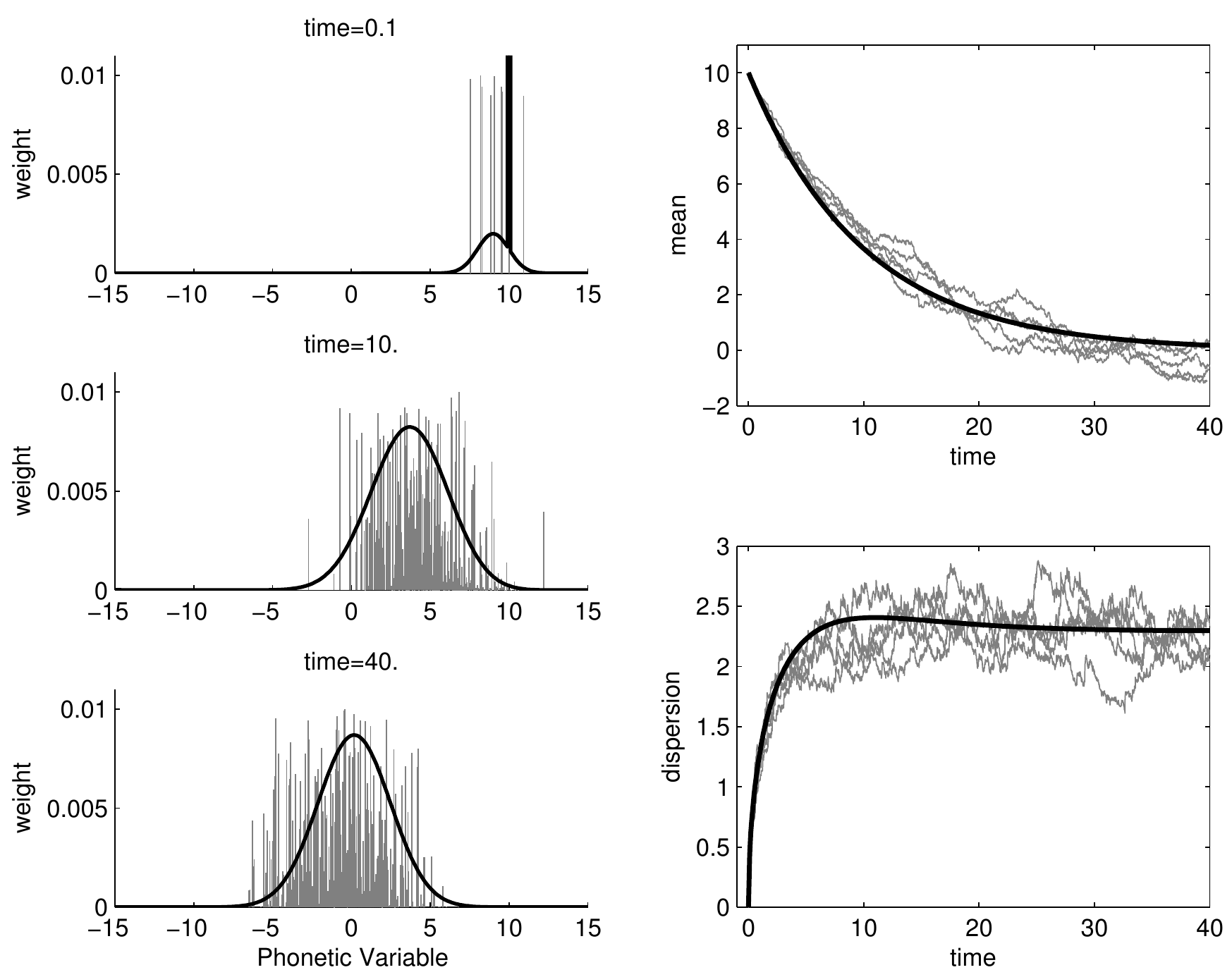}
\end{center}
\caption{  \label{fig:one_category_exemplars} In grey, exemplar model, one category, one dimension. $\nu=100$, $\lambda=1$, $W_0=1/\nu$, $\alpha=0$, $\beta=0.1$, $y^*=0$. In black, the corresponding field model.  On left, exemplars positions and weights after 0.1, 10, and 40 time units, along with $\rho$ for the field model. ($\rho$ was rescaled by a factor of $1/20$ for ease of comparison.)  On right, weighted mean $\bar{y}$ and dispersion $\nu$ plotted against time for  five different trials, along with corresponding values for the field model.}
\end{figure}

In Figure~\ref{fig:one_category_exemplars} we show the results of a simulation of this model. We do not intend the variable $y$ to have any particular phonetic interpretation and we have chosen $y^*=0$ for convenience. The grey lines indicate results from the exemplar model, and the black lines indicate results from the field model we will derive in the next subsection. On the left we show the location and weight of all the exemplars in the system at three points in time. On the top right we show how the exemplar  mean $\bar{y}$  varies with time for five different simulations of the exemplar model. 
The mean converges to nearby $y^*$ in each simulation.
On the bottom right, again for five different simulations, we show how
 the \emph{dispersion} of the exemplars varies with time, where dispersion defined by
\[
s : = \sqrt{ \frac{\sum_{i=1}^n w_i (y_i-\bar{y})^2  }{\sum_{i=1}^n w_i}},
\]
is the weighted standard deviation of $y_i$ about $\bar{y}$. Dispersion gives a measure of the width of the distribution of exemplars in phonetic space.

\subsection{Derivation of the Field Model: Single Word} \label{subsec:onecat_nonloc}

Here we derive the exemplar density field model for the case of a single word, where the phonetic value of the exemplars are represented by a single scalar.  The exemplar density field is denoted $\rho(y,t)$ and gives the weighted density of exemplars at the point $y$ in phonetic space and at time $t$. We consider a limit in which the rate of production of exemplars $\nu$ goes to infinity, and the initial weight of each new exemplar $W_0$ goes to 0.  Here we will derive evolution equations for $\rho(y,t)$ in this case, and demonstrate some of its simple properties.

In the previous subsection, the state of the system at time $t$ consisted of $n(t)$ exemplars with positions $y_i$ and weights $w_i$.  We define the exemplar density to be
\[
\rho(t,y) = \sum_{i=1}^n \delta(y-y_i)w_i(t),
\]
where $\delta$ is the $\delta$-distribution.
Defined this way, for all time the field will consist of a finite number of $\delta$ peaks.
 We consider the weak limit of this  field as the rate of exemplar production $\nu$ goes to infinity  and the initial exemplar weight $W_0$ goes to zero. We derive an evolution equation for the field $\rho$ in this  limit.
  We assume that the distribution of exemplar weights and exemplar positions are approximately independent of each other so that we can write a closed expression for this field. Later we will validate this assumption by comparing simulations of the exemplar model and the  field model in  Subsection~\ref{subsec:numer}.

Recall that new exemplars are produced at rate $\nu$ and that  the weight of each exemplar decays at rate $\lambda$.  
If each exemplar is created with weight $W_0$ then we must have $W_0 \nu=\lambda W$ at equilibrium, where $W$ is the total weight in equilibrium.  We fix $\lambda$ and let $W_0$ go to zero while $\nu$ goes to infinity.
The parameters $\nu$ and $W_0$ should be inversely related to guarantee a fixed $W$ limit, so we let $W_0=\mu/\nu$, where $\mu$ is a constant. We derive equations for the exemplar density field in this limit.

Consider the interval $[y, y + \Delta y]$.  We indicate the integral of $\rho$ over the interval by $R$.
We estimate the mean rate of change of $R$.
Each exemplar's weight is decaying at rate $\lambda$.  So the rate of  change of $R$ with respect to time due to decay of exemplars is 
$-\lambda R$. 
New exemplars may be produced in the interval by any of the other exemplars. The rate of exemplar production in the interval $[y,y+\Delta y]$ due to exemplars in an interval $[z , z+ \Delta z]$ is, to leading order,
\[
\frac{\nu \, \Delta y\, \Delta z\, f(y-z - \alpha (\bar{y}-z) - \beta (y^* - z) )\, \rho(z) }{\int \rho(z) dz}
\]
(dropping the $t$ dependence)
where $f$ is the density of Gaussian with mean zero and variance $\sigma^2$.
Multiplying by $W_0$ (to get the rate of exemplar \emph{weight} production), letting $\Delta z \rightarrow 0$, and integrating over all $z$ gives the rate of exemplar production to be
\[
 \mu \Delta y \frac{ \int  f(y -z  - \alpha(\bar{y}-z) - \beta( y^* - z) ) \rho(z) dz }{ \int \rho(z) dz}.
\]
This yields 
\[
\frac{dR}{dt} = - \lambda R +  \mu \Delta y \frac{ \int  f(y -z  - \alpha(\bar{y}-z)  - \beta (y^* - z) ) \rho(z) dz }{ \int \rho(z) dz}.
\]
Finally, dividing by $\Delta y$ and letting $\Delta y \rightarrow 0$ gives 
\begin{equation}\label{eq:onecat_nonloc}
\frac{\partial \rho(y)}{\partial t} = - \lambda \rho(y) +  \mu \frac{\int f(y- z - \alpha (\bar{y}-z) - \beta (y^* -z) ) \rho(z) dz }{\int \rho(z) dz},
\end{equation}
which is the evolution equation for the exemplar density field $\rho$. This equation is what we will be generalizing to the case of multiple words in Subsection~\ref{subsec:marmot2}.

We can analytically derive some properties of solutions of \eqref{eq:onecat_nonloc}  
We define the \emph{total weight}
\[
M(t) := \int \rho(t,y) \, dy
\]
to be the integrated weight of all exemplars. If we integrate \eqref{eq:onecat_nonloc} with respect to $y$ we obtain
\[
\frac{dM}{dt} = - \lambda M + \mu
\]
showing that the total weight converges to an equilibrium value of $M=\mu/\lambda$.

The analogue of exemplar mean for the field model is 
\[
\bar{y} = \frac{ \int y \rho(y) \, dy}{\int \rho(y) \, dy} =   \frac{ \int y \rho(y) \, dy}{M}.
\]
Multiplying  \eqref{eq:onecat_nonloc} by $y$ and integrating with respect to $y$  gives
\[
\frac{d\bar{y}}{dt} = - \frac{\mu \beta}{M} (\bar{y}- y^*).
\]
So $\bar{y}$ converges to the preferred phonetic value $y^*$ assuming $\beta>0$.

To compute the equilibrium for $\rho(y)$, we set $\partial \rho/\partial t$ to zero and assume that $M=\mu/\lambda$ and $\bar{y}=y^*$. This gives
\[
\rho(y) = \int f(y-z + (\alpha +\beta) z ) \,  \rho(z) \, dz
\]
Rewriting the expression on the right as a convolution by doing a change of variables in $z$, taking the Fourier transform and solving for $\rho$ gives a Gaussian equilibrium
\begin{equation}  \label{eqn:gaussian}
\rho(y) = \frac{\mu}{\lambda} \frac{1}{\sqrt{2 \pi s^2}} e^{ - (y-y^*)^2/ 2 s^2} 
\end{equation}
where the dispersion $s$ is given by 
\begin{equation} \label{eqn:gauss_dispersion}
s^2 : = \frac{ \int (y-\bar{y})^2 \rho(y) \, dy}{\int \rho(y) \, dy}  = \frac{\sigma^2}{ 1 - (1 - (\alpha+\beta)) ^2}.
\end{equation}
Note that $\alpha$ and $\beta$ do not have independent effects on the equilibrium:
even if $\alpha=0$ (no entrenchment), the category will be stably located around $y=y^*$ as long as $\beta>0$ (lenition).
In Figure~\ref{fig:one_category_exemplars} the black lines show the results of
simulation of this field model, which we see captures many
of the qualitative features of the exemplar model.

\section{Exemplar Dynamics and Field Models for Multiple Words} \label{sec:multiple}

In this section we generalize the models of Section~\ref{sec:onecategory} to describe the interactions between exemplars of multiple words.  We continue to represent exemplars with a single phonetic variable.  We develop models for the case of two interacting words here; the extension to more than two is straightforward.

For the multiple word exemplar model in Subsection~\ref{subsec:onecat_exemp}, production and decay of exemplars proceeds as before, but after a new exemplar is produced it is assigned to one of the existing words according to a categorization procedure. In particular, we describe the three \emph{categorization regimes} that we study. The choice of categorization regime will turn out to determine whether or not the pronunciation of two words merges or not.

In Subsection~\ref{subsec:marmot2} we derive the field model limit of the exemplar model in Subsection~\ref{subsec:marmot1}. This is the two-word analogue of the equation for $\rho$ given in \eqref{eq:onecat_nonloc}.

\subsection{Exemplar-Level Model: Multiple Words} \label{subsec:marmot1}

Each exemplar now belongs to one of two words, $A$ or $B$. Exemplars are produced from each words with rates $\nu_A$, $\nu_B$ respectively. We assume that all exemplars are created with initial weight $W_0$.  The weights of the exemplars decay with rate $\lambda$ as before.  

To describe how categorization works, suppose a new exemplar is produced from word $A$.  An existing exemplar from word $A$ is selected randomly with probability proportional to exemplar weight. As in the one-word case, if $z$ is the position of the existing exemplar, a new exemplar is  produced at 
\[
y= z + \alpha (\bar{y}_A - z)  +  \beta (y^*- z )+ \sigma \eta
\]
where $\bar{y}_A$ is the weighted mean of the positions of the exemplars from word $A$, and $y^*$ is a  preferred location in phonetic space as before.

The new exemplar has to be classified as either belonging to word $A$, classified as belonging to word $B$, or discarded. We consider three different ways that this may occur, which we call  {\emph{Categorization Regimes}}. We first describe the Categorization Regimes at a high level, and then give details.

\vspace{1cm}
{\bf Categorization Regimes}
\vspace{0cm}
\begin{enumerate}
\item {\emph{No Competition:}} Exemplars are always classified into the word they were produced from.
\item {\emph{Pure Competition:}} Words compete to claim the new exemplar as their own, without information about the word it came from. The winning word accepts the exemplar into its collection of exemplars.
\item {\emph{Competition with Discards:}} Words compete to claim the new exemplar as their own, without information about the word it came from. If the exemplar is claimed by the word it came from, then it is stored in that word. If the exemplar is claimed by a different word then it is discarded.
\end{enumerate}
\vspace{.3cm}

In the first, the \emph{no-competition} regime, the exemplar is assigned to the word that it was produced from. Linguistically, this means that the speaker is able to determine from context what word was intended and store the exemplar as an instance of  that word, regardless of the actual sound produced by the speaker. In the no-competition regime there is no interaction between exemplars of different words. So the model consists of the exemplars for word $A$ and the exemplars for word $B$ both evolving independently of one another.

 In the second, the \emph{pure competition} regime, the hearer classifies an exemplar based on its phonetic variables: the words ``compete" to acquire the exemplar.
The idea is that the  new exemplar is more likely to be classified as an instance of word  $A$ if there are more exemplars of word $A$ with strong weight  close to the new exemplar in phonetic space.  
We assume that the speaker's word that the exemplar was selected from  does not figure in the assignment. This is the mechanism for
prevention of sound merger proposed in \cite{blevins_wedel}.

In the third, the \emph{competition with discards} regime, the hearer decides how to classify the exemplar exactly as in the pure competition regime. But when the exemplar is misclassified (i.e.\ assigned to a different word than where it was generated from) the hearer is able to identify this fact and the exemplar is discarded rather than being stored. The hearer may do this by one of several means.  For example, if the word resulting from the misclassification does not exist, or if the resulting sentence is non-sensical, this indicates to the speaker that there has been a misclassification.

To give an example of the three models at work within a linguistic context, consider the following example in which the two relevant words are the English ``pin" and ``pen". A speaker wants to give a phone number to the hearer and intends to utter the sentence 
\begin{center}
``You have a pen?"
\end{center}
but what is received by the hearer is
\begin{center}
``You have a pXn?"
\end{center}
where $X$ is some indeterminate vowel sound between an unambiguous `i' and an unambiguous `e'.  We assume that `X' is closer to `i' than to `e'.

\emph{No Competition.} The hearer classifies `pXn' as an example of  `pen', because `pen' is obviously what is intended in this context.
 Thereafter `pXn' is stored as an exemplar of `pen'.

\emph{Pure Competition.} Since `X' is closer to an `i' than an `e', `pXn' is classified as `pin'. Thereafter, `pXn' is stored as an exemplar of `pin'. The meaning of the sentence plays no role in the classification.

\emph{Competition with discards.}  Since `X' is closer to an `i' than an `e', `pXn' is classified as `pin'. However, since the sentence doesn't make any sense in this context with `pin',  `pXn' is \emph{not} stored as an exemplar of the word `pin'. The sound `pXn' is discarded and not stored as an exemplar of any word.

In the two competitive regimes we need an explicit model of how the words compete for an exemplar.
In order for the classification of a new exemplar to be influenced by the weight and number of nearby exemplars  of both words, we define local smoothed exemplar density fields:
\begin{equation} \label{eqn:hyrax1}
S_A(y)  =  \sum_{i=1}^{n_A} \frac{1}{2} k w_{A,i} e^{-k |y-y_{A,i}|}, \ \ \ \
 S_B(y)  =  \sum_{i=1}^{n_B} \frac{1}{2} k w_{B,i}  e^{-k |y-y_{B,i}|}.
\end{equation}
The factor of $\frac{1}{2}k$ guarantees that $\int_y S_A(y) dy = \sum_i w_{A,i}$ and likewise for word $B$. As $k \rightarrow 0$ the smoothed fields are uniform in space.  As $k \rightarrow \infty$ there is no smoothing and $S_A(y) = \rho_A(y)$, $S_B(y)=\rho_B(y)$.
We define the  probability of the new exemplar generated at point $y$ being assigned to word $A$ or $B$ as  
\begin{equation*}
f_A(y) =  \frac{S_A(y)^p}{S_A(y)^p +S_B(y)^p}, \ \ \ \ \ f_B(y) =  \frac{S_B(y)^p}{S_A(y)^p +S_B(y)^p},
\end{equation*}
respectively,
where $p \geq 0$ is a fixed selection parameter. For $p=1$ the exemplar is assigned to a new word in direct proportion to the magnitude of the smoothed field at $y$.  In the  $p\rightarrow \infty$ limit the exemplar is assigned to whichever word has greater smoothed density, $S_A$ or $S_B$, at $y$.

\subsection{Derivation of the Field Model: Multiple Words} \label{subsec:marmot2}
 
 Taking the same limit as we do in the single-word case we can derive equations for the exemplar density fields of multiple words.
 As in the single word case, we define the exemplar density fields by
 \[
 \rho_A(y) = \sum_{i=1}^n \delta(y-y_{A,i}) w_{A,i}(t)
 \]
 and similarly for $\rho_B$. We let $\nu_A$ and $\nu_B$ go to infinity as $W_0 \rightarrow 0$, with 
 $\mu_A= \nu_A W_0$ and $\mu_B = \nu_B W_0$.
 We define the analogue of the smooth density field for the field model using convolution with a smoothing kernel.
 Let $K$ be the smoothing operator given by
 \begin{equation} \label{eqn:hyrax2}
 {K}\rho_A(y) = \int_z \frac{1}{2} k \rho_A(z)  e^{-k|z-y|} \, dz. 
 \end{equation}
 Then ${K}\rho_A$ and ${K}\rho_B$ are the smoothed density fields.
  The probability of a new exemplar generated at $y$ are being classified as word $A$ or $B$ are respectively
\[
 f_A(y) =  \frac{[{K}\rho_A(y)]^p}{[{K}\rho_A(y)]^p +[{K}\rho_B(y)]^p}, 
 \ \ 
 f_B(y) =  \frac{[{K}\rho_B(y)]^p}{[{K}\rho_A(y)]^p +[{K}\rho_B(y)]^p}. 
\]
 
 We define the field  production terms by
 \[
 P_A(t,y) = \mu_A \frac{\int f(y- z - \alpha (\bar{y}_A-z) - \beta (y^*-z) ) \rho_A(z) dz }{\int \rho_A(z) dz}
 \]
 and the corresponding expression for $P_B$.
 
The equations for the no-competition regime are then
\begin{equation} \label{eqn:noncomp}
\begin{aligned}
\frac{\partial \rho_A(y)}{\partial t} & =  - \lambda \rho_A(y) +   
P_A(t,y),  \\
\frac{\partial \rho_B(y)}{\partial t} & =  - \lambda \rho_B(y) +  
 P_B(t,y). 
 \end{aligned}
\end{equation}
We can see that two fields evolve independently of each other.
 
 The equations for the pure competition regime are 
\begin{eqnarray} \label{eqn:purecomp}
\begin{aligned}
\frac{\partial \rho_A(y)}{\partial t} & =   - \lambda \rho_A(y) +  f_A(y) \left\{ 
P_A(t,y) + P_B(t,y)
\right\}, \\
\frac{\partial \rho_B(y)}{\partial t} & =   - \lambda \rho_B(y) +  f_B(y) \left\{ 
P_A(t,y) + P_B(t,y)
\right\}.
\end{aligned}
\end{eqnarray}
 Here we see that it is only the \emph{sum} of $P_A$ and $P_B$ that matter to the dynamics of the fields and not their individual matters.
 
 Finally, the equations for the competition with discards regime are
 \begin{equation} \label{eqn:compdisc}
 \begin{aligned}
\frac{\partial \rho_A(y)}{\partial t} & =  - \lambda \rho_A(y) +  f_A(y)
P_A(t,y), \\
\frac{\partial \rho_B(y)}{\partial t} & =  - \lambda \rho_B(y) +  f_B(y)  P_B(t,y). 
\end{aligned}
\end{equation}
The production term $P_A$ only contributes directly to the evolution of $\rho_A$, and likewise for $B$.

\subsection{Properties of the Field Models}
\label{subsec:analprop}

Using the no-competition regime the system is just two uncoupled versions of the single word case. 
Both $\rho_A$ and $\rho_B$  converge to the equilibrium Gaussian given by \eqref{eqn:gaussian} and \eqref{eqn:gauss_dispersion}. Thus the words merge in this regime. 

Once we introduce competition, analytically we only have a few simple results, and only for the pure competition regime.
 Adding the equations \eqref{eqn:purecomp} for $\rho_A$ and $\rho_B$ together gives
\[
\frac{\partial \rho(y)}{\partial t} = - \lambda \rho(y) + P_A(y)  + P_B(y),
\]
where $\rho= \rho_A + \rho_B$.
 Integrating over the whole domain with respect to $y$ gives
 \[
 \frac{dM}{dt} = - \lambda M + \mu_A + \mu_B.
 \]
 for which we find the equilibrium value $M=(\mu_A + \mu_B)/\lambda$ for the total weight of both words.
  
 We can say even more in the special symmetric case when $\mu_A=\mu_B=\mu$, $y^*=0$,
$\rho_A(y) = \rho_B(-y)$ and $\alpha=0$.  The condition on $\rho_A$ and $\rho_B$ is invariant under the flow of the dynamics and implies that $y_B= - y_A$ and 
 $M_A = M_B =M/2$.
  Under these conditions we have
   \[
 P_A(y) + P_B(y) = \frac{2\mu}{M}   \int f(y-z + \beta z ) \rho(z) dz.
 \] 
 So in this symmetric case the equation for the total field is
 \[
 \frac{\partial \rho(y)}{\partial t} = - \lambda \rho(y) + \frac{2\mu}{M}   \int f(y-z + \beta z ) \rho(z) dz.
 \]
 This is identical in form to the equation for $\rho$ in the single word model. So we know that there is a Gaussian equilibrium  $\rho$, but this does not give us an equilibrium for $\rho_A$ and $\rho_B$, which will depend on other parameters including the selection parameter $p$.

\subsection{Numerical Comparison of Exemplar Model and Field Model} \label{subsec:numer}

Here we explore  the behaviour of both our exemplar models and the  field models under different categorization regimes. We also validate the field models as an approximation to the exemplar models, showing that they closely agree when the rate of exemplar production is sufficiently large.

In all of the following simulations we choose $\alpha=0$, meaning that there is no bias in production towards the mean phonetic value of the  exemplars. (Having $\alpha>0$ does not qualitatively change the results.)  In each case $\beta= 0.1$, $\lambda=1$, $y^*=0$, $k=10$, and $W_0=1/\nu$, leading to $\mu=1$. We vary the categorization regime and the selection parameter $p$. 

\begin{figure}
\begin{center}
\includegraphics[width=12cm]{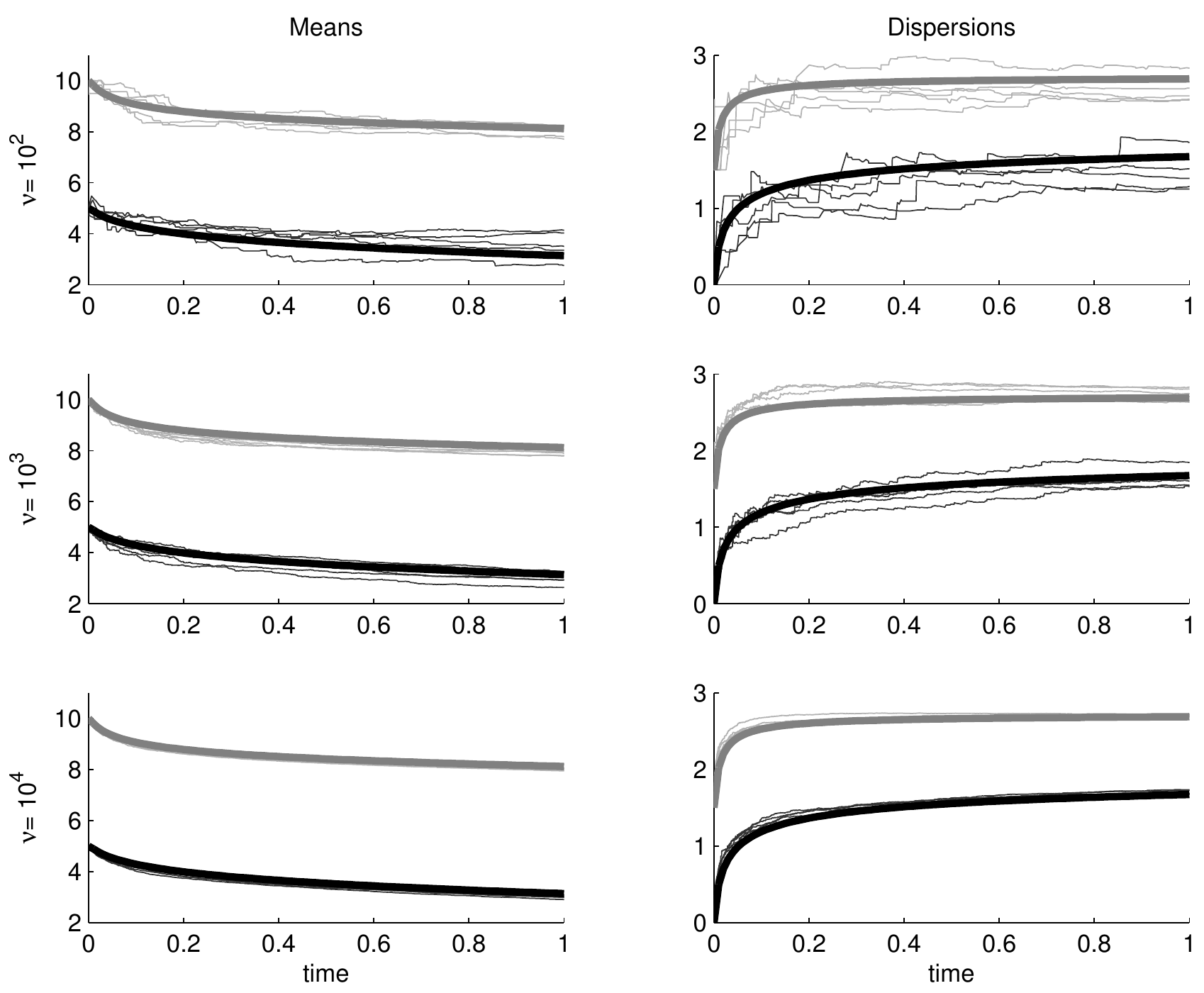}
\end{center}
\caption{\label{fig:validateRegime2} Simulation of two words interacting in the  competition with discards regime with the selection parameter $p=1$.  Exemplar means and dispersions versus time are shown for five different simulations of the exemplar model (thin lines) and the field model (thick lines). Word A (black) and word B (grey). Top: $\nu=10^2$, Middle: $\nu=10^3$, Bottom: $\nu=10^4$. On the right, the dispersions for word B are offset by 1.5 unit for clarity.}
\end{figure}

First we validate the field model as given in \eqref{eq:onecat_nonloc} as an approximation to the exemplar system in the limit of $\nu$ going to infinity. We present this for the case of the competition with discards regime with selection parameter $p=1$.
It is only computationally feasible to perform this validation on short time intervals, since the computational cost of the exemplar model grows considerably as $\nu$ grows.
In Figure~\ref{fig:validateRegime2} we plot exemplar means and dispersions for five simulations of the exemplar model and one simulation of the field model.
We see that as $\nu$ goes to infinity the trajectories of the exemplar simulations converge on the field simulation.
We obtained similar convergence results for the other regimes but we do not present them here.

\begin{figure}[h!]
\begin{center}
\includegraphics[width=6.18cm]{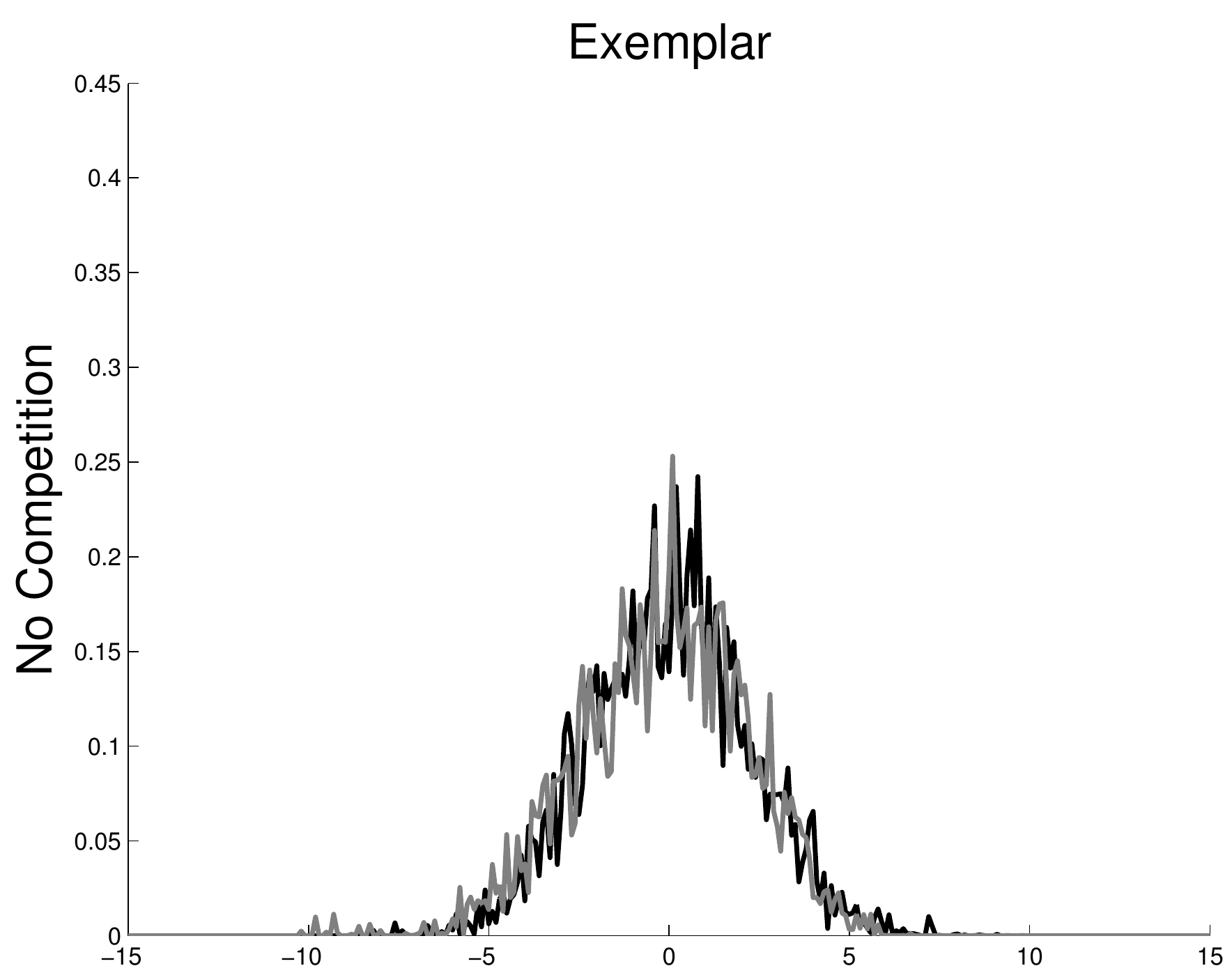}
\includegraphics[width=5.82cm]{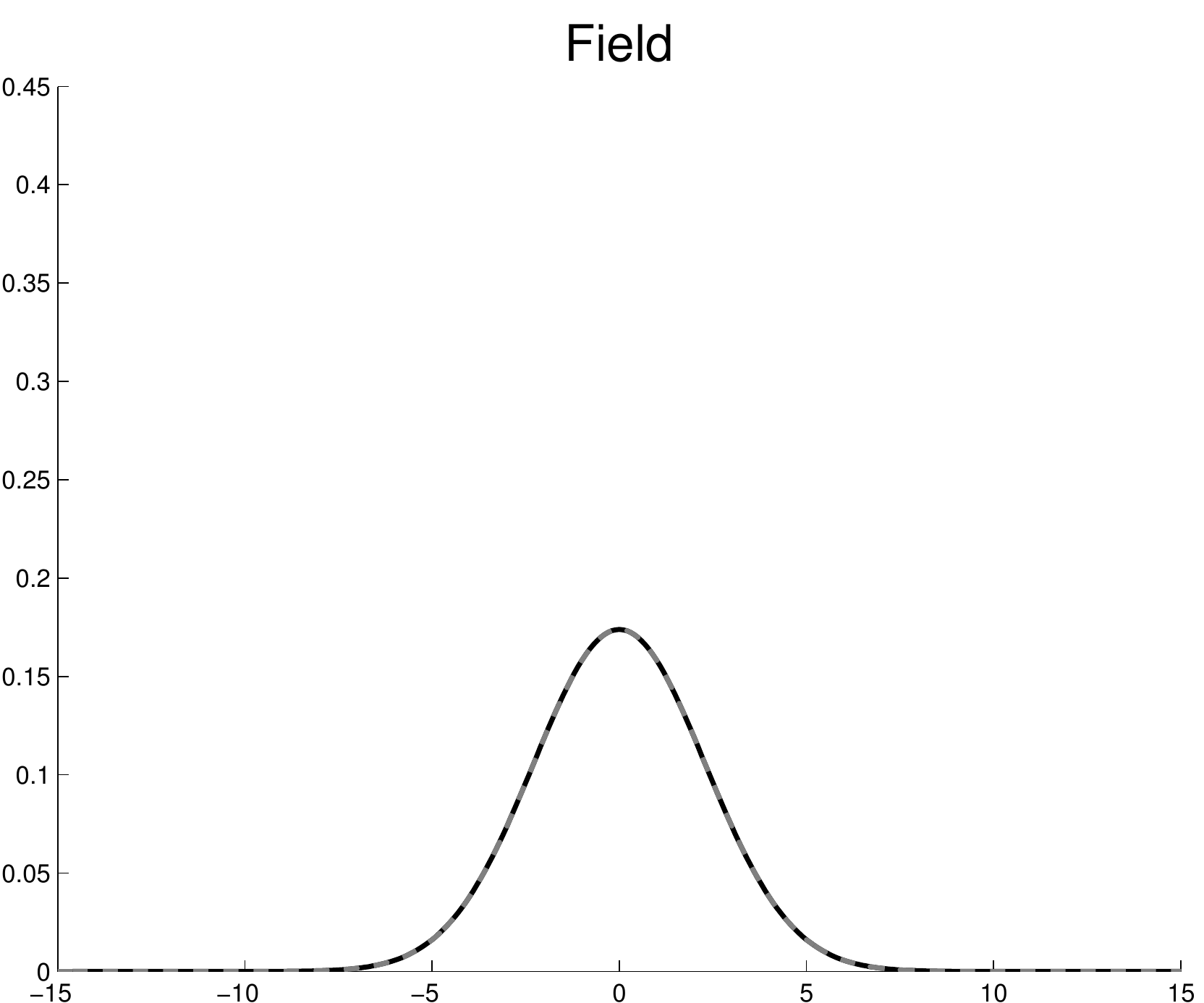} \\
\includegraphics[width=6.18cm]{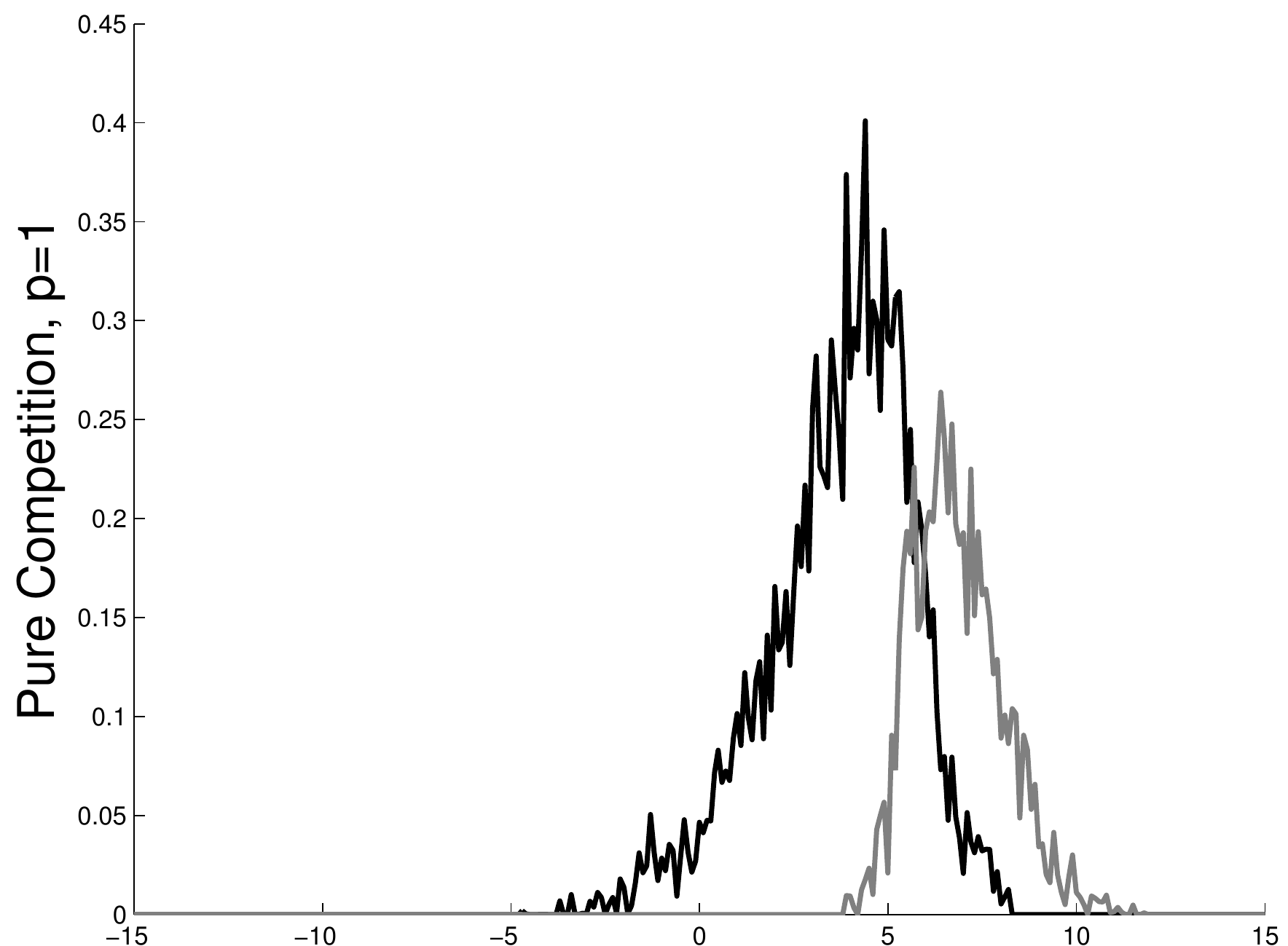}
\includegraphics[width=5.82cm]{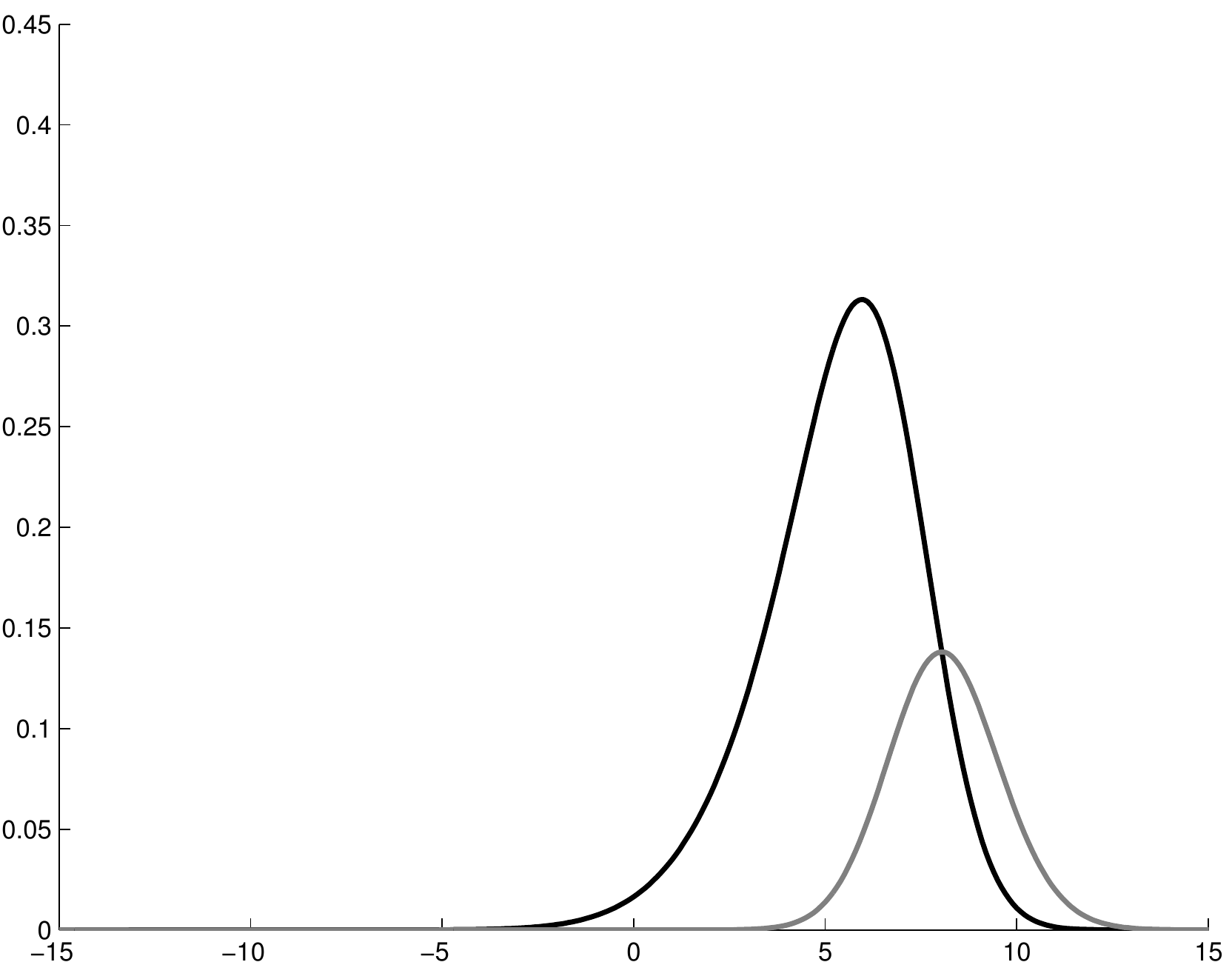} \\
\includegraphics[width=6.18cm]{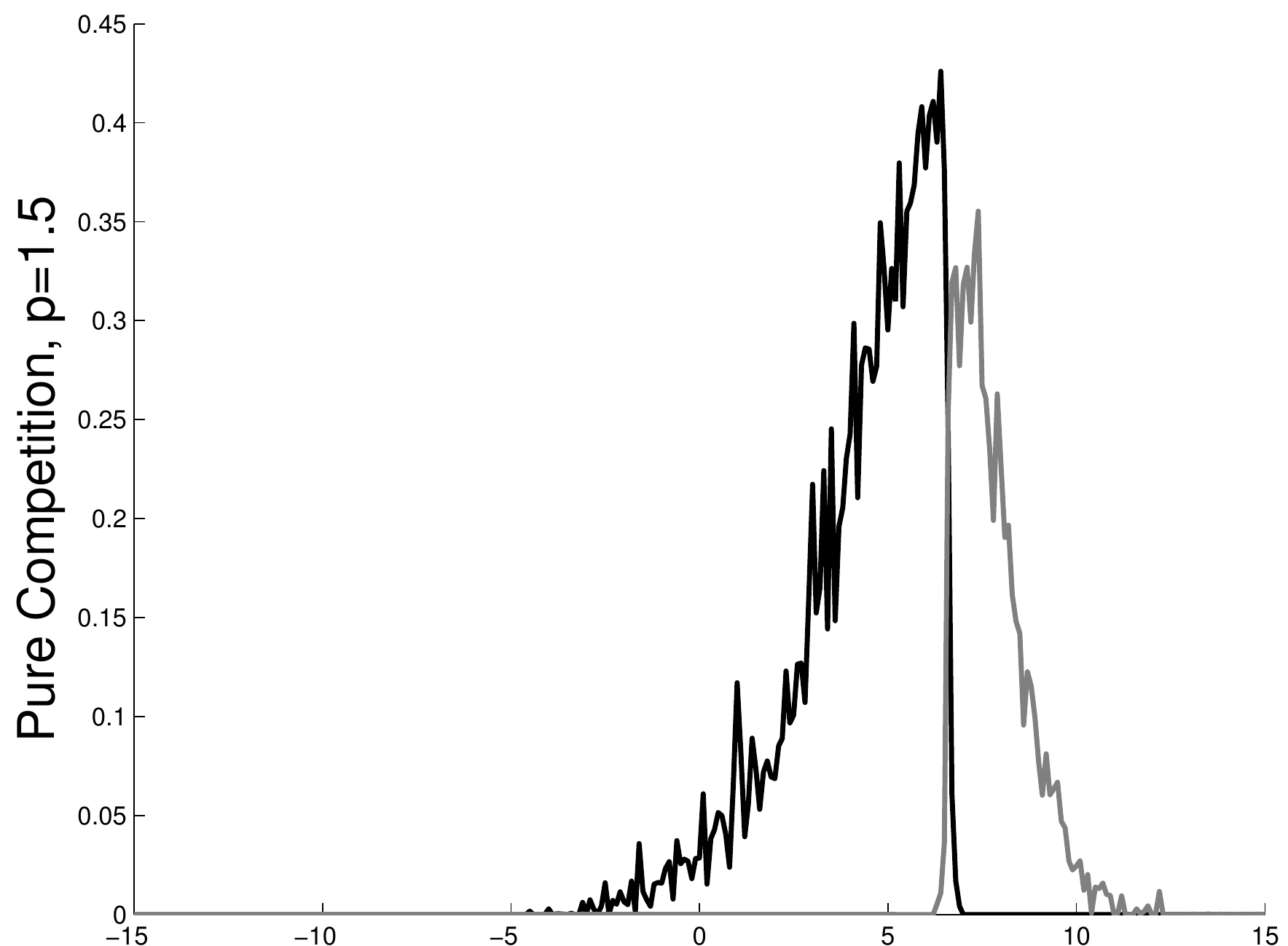}
\includegraphics[width=5.82cm]{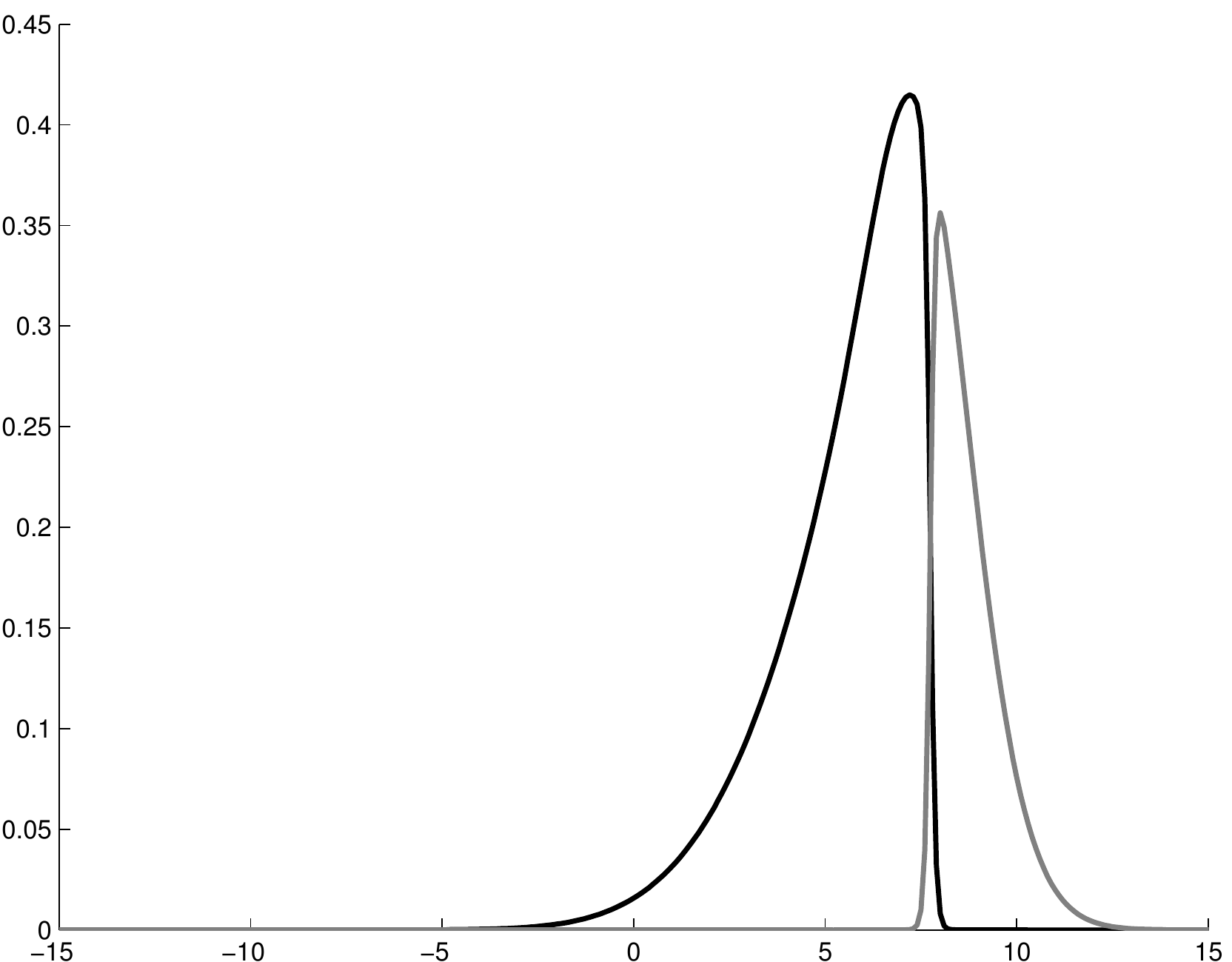} \\
\includegraphics[width=6.18cm]{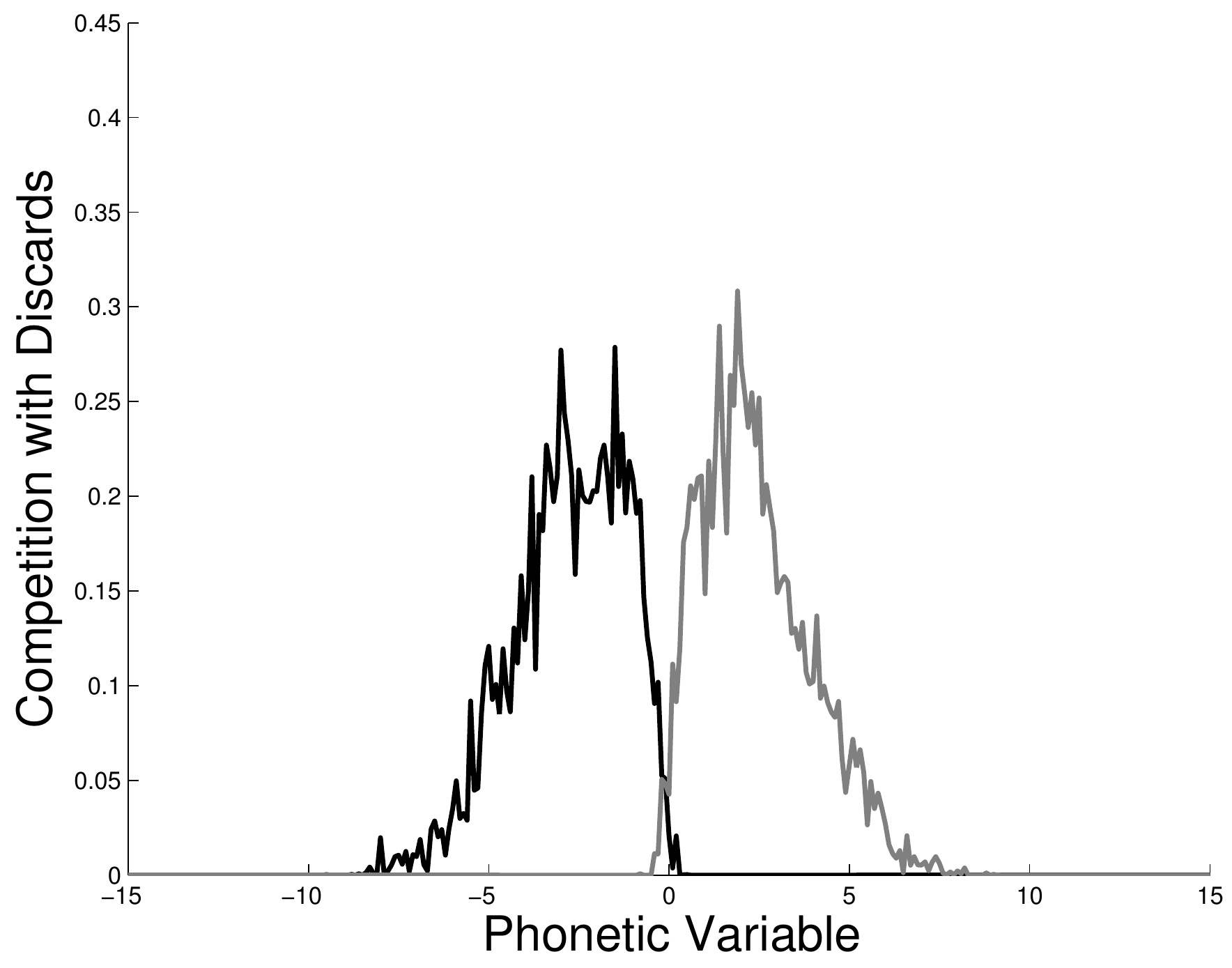}
\includegraphics[width=5.82cm]{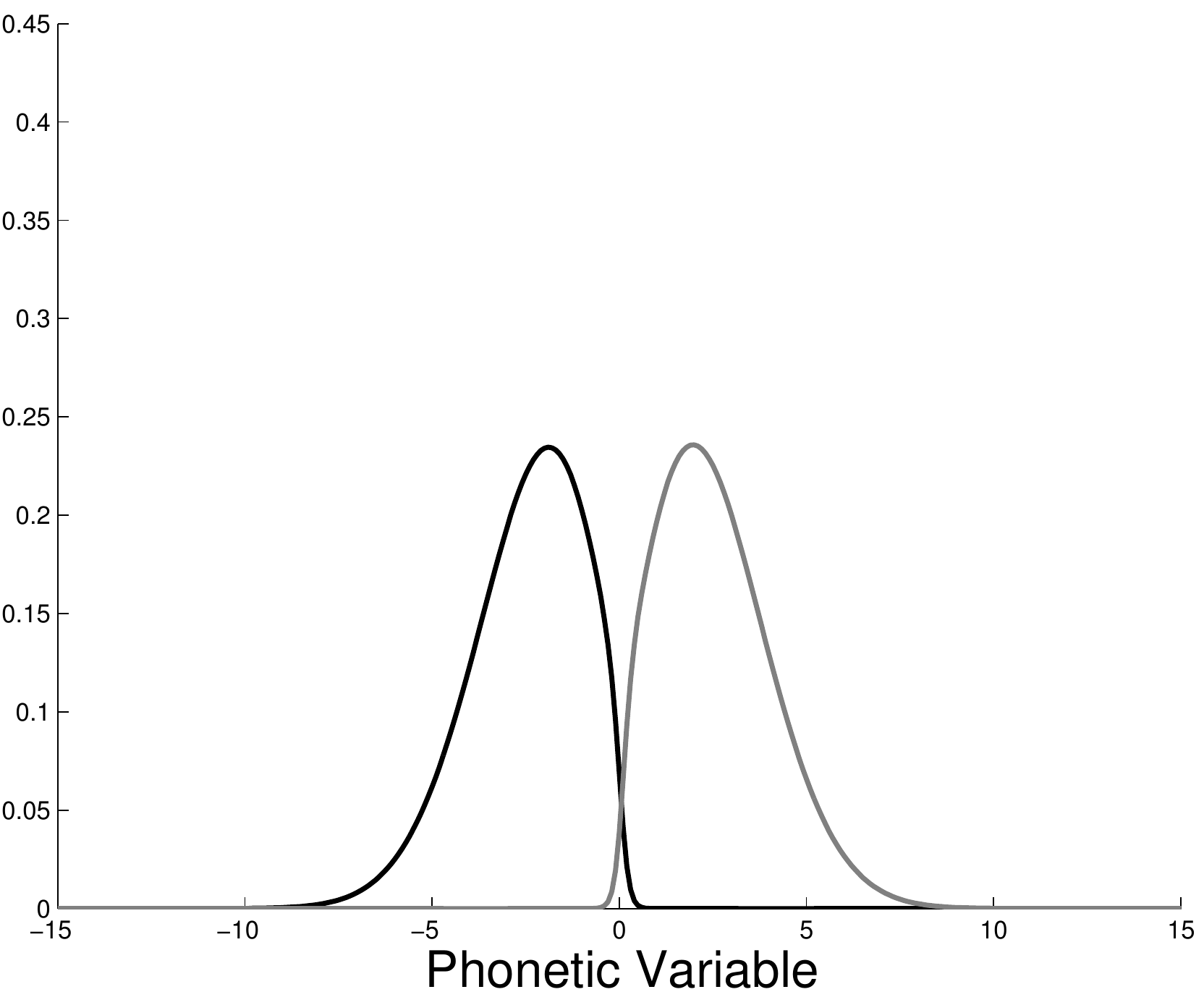}
\end{center}
\caption{ \label{fig:blah} Exemplar density fields $\rho_A$ (black) and $\rho_B$ (grey) for the exemplar model (left) and the field model (right) at $t=100$. Four different categorization regimes/choice of selection parameter $p$ are shown. }
\end{figure}

 In order to explore the different regimes we performed simulations of both the exemplar model and the field model over 100 time units. For the exemplar model we use $\nu=10^3$ and  initialized the system with $50$ exemplars for word $A$ at $y=5$ and $50$ exemplars for word $B$ at $y=10$. The field model is initialized with a matching initial condition. The results of these simulations are presented in Figure~\ref{fig:blah}.  We show a plot of $\rho_A$ and $\rho_B$ at time $t=100$, with the exemplar model on the left and the corresponding field model on the right.
  We also present in Figure~\ref{fig:long}  results of a longer simulation  (over 1000 time units) of the field model alone. In this latter figure we show $\rho_A$ and $\rho_B$ at time $t=1000$ in the first column, and in the remaining three columns we plot  means,  dispersions, and total weight of words  versus time. We discuss each of the different choices of categorization regime and selection parameter $p$ in turn.

\subsubsection{No Competition:}
Using the no-competition regime the system is just two uncoupled versions of the single word case. 
Both $\rho_A$ and $\rho_B$  rapidly converge to the equilibrium Gaussian of the single word case, as can be seen in the top row of Figure~\ref{fig:blah} and  in the top row of Figure~\ref{fig:long}.
The pronunciation of the two words has merged.

\subsubsection{Pure Competition: $p=1$} 
We show the results for this regime in the second rows of Figure~\ref{fig:blah} and Figure~\ref{fig:long}. 
The plots demonstrate that even though the pronunciations of the two words start off separated, they slowly merge over the course of the simulation.  This impression is confirmed by a longer simulation of the field model which is shown in the second row of Figure~\ref{fig:long}. At $t=1000$ the exemplars of the words have nearly the same means, and the means appear to be converging to $\rho^*=0$.

We argue that the complete merger for this case is due to choice of the selection parameter $p=1$.   To see this, imagine the system in a state where there is a region of phonetic space with $2/3$ of the weight due to exemplars of word $A$ and $1/3$ of the exemplars of word $B$.  When $p=1$, a new exemplar produced in the location has probability $2/3$ of being classified as $A$ and $1/3$ of being classified as $B$.
So the expected change in the ratio between the weights of the two words is zero. Neglecting variation of $\rho_A$ and $\rho_B$ with respect to $y$, coexistence exemplars of two words in any admixture is stable.

We conclude that, even though it may be a realistic model of categorization in some contexts,  the pure competition regime with selection parameter $p=1$ cannot account for the existence of  stable distinct pronunciations in phonological systems.

\subsubsection{Pure Competition: $p=1.5$}
\begin{figure}
\begin{center}
\includegraphics[width=13cm]{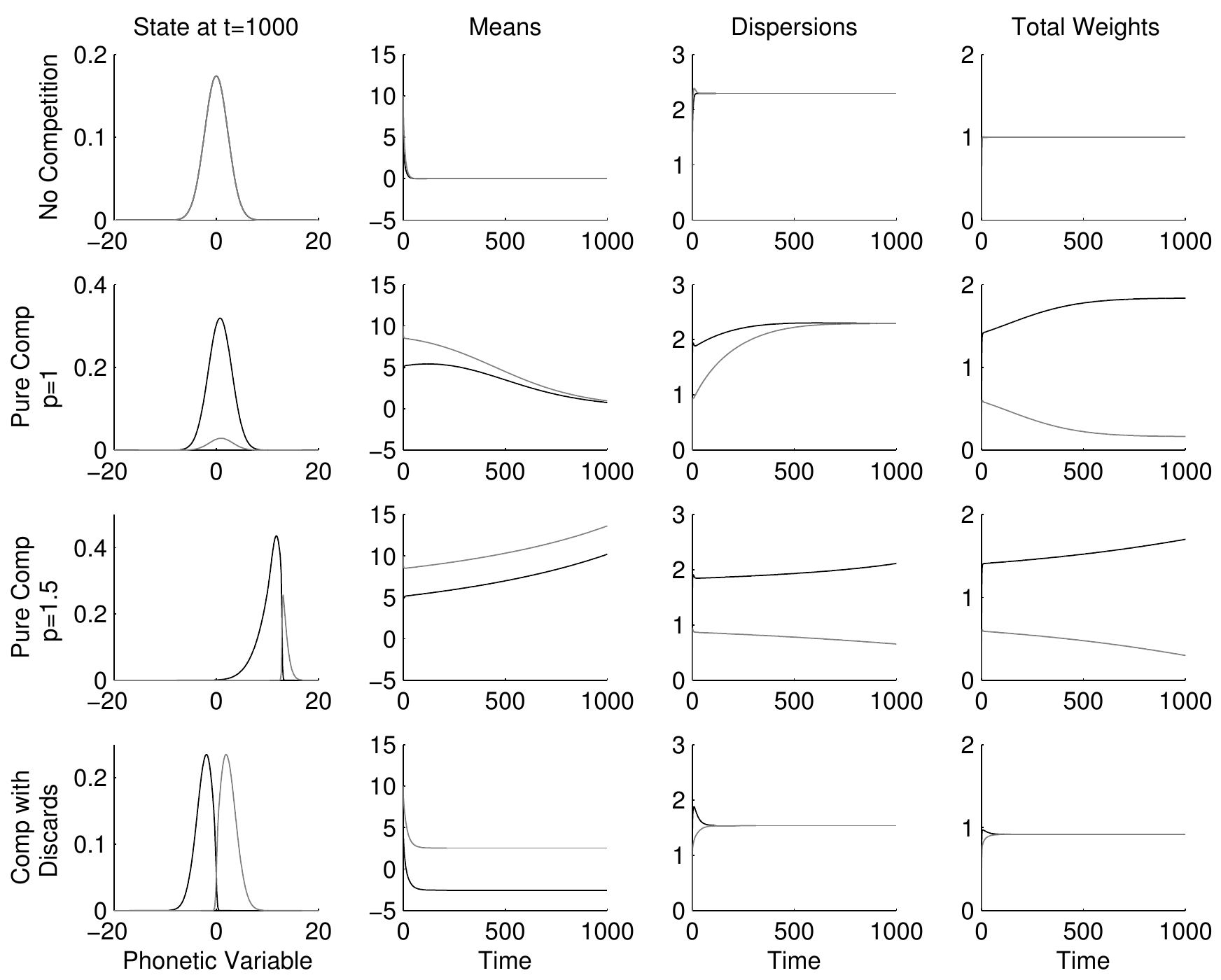}
\end{center}
\caption{\label{fig:long} Each row shows results for the field simulation with a different categorization regime.
The first column shows the exemplar density $\rho$ as a function of $y$ at time $t=1000$. The three remaining columns show means, dispersions, and total weight of each word versus time.
}
\end{figure}

The argument just made for why the pure competition regime leads to sound merger does not hold when the selection parameter $p>1$. Indeed, with this choice, the more active word in a region of phonetic space draws in more than its share of new exemplars. To test this idea, we perform the same simulation as we did for $p=1$ for $p=1.5$. The results are shown in the third rows of  Figure~\ref{fig:blah} and Figure~\ref{fig:long}.
 
 As predicted, the words  do not merge and appear to remain sharply distinct for all time. However, there is another anomaly: the system does not appear to settle into an equilibrium that is symmetric about $y^*=0$.
In Figure~\ref{fig:blah} it seems that the two exemplar clouds have a boundary near $y=8$ at time $t=100$. 
 Looking at the longer simulation of the field model in  the third row of Figure~\ref{fig:long} we see that in fact the exemplar clouds are slowly moving to the right.

Here is an explanation for the system's behaviour in this situation. With the choice $\nu_A=\nu_B$ the two words are selected equally often for production of a new exemplar. However, due to the bias towards $y^*=0$ in production, new exemplars produced from the word on the right ($B$) are more likely to be categorized as belonging to the word on the left ($A$) than vice versa. 
This has the result of shifting the means of the exemplars of both word $A$ and word $B$ to the right, as well as increasing the total weight of word $A$ over that of  word $B$. The increased weight of word $A$  only increases the rate at which exemplars from word $B$ are classified as word $A$, shifting word $A$ to the right even further. This feedback loop continues with greater disparity in weight  leading to more classification of exemplar from $B$ into $A$, and an even stronger push towards greater $y$ values.   Evidently, this effect is strong enough to counteract the effect of the bias in production towards $y^*=0$.
 We conjecture that both words continue to move to the right indefinitely.

Again, the pure competition regime with $p=1.5$ may be a realistic model in some circumstances. However, it is unlikely to be a very general mechanism for the stability of phonological distinctions, since even though the pronunciations of the words do not merge, they are not stably located within phonetic space.

\emph{Competition with Discards. $p=1$.} Finally we consider competition with discards regime. Choosing selection parameter $p=1$, we plot the state at time $t=100$ in the bottom row of Figure~\ref{fig:blah}. (Other choices of selection parameter $p\geq 1$ do not  change the results qualitatively.) Here the system approaches a equilibrium where the exemplars of the two words occupy distinct portions of phonetic space and they are symmetric about $y^*=0$.  This is confirmed by the longer simulation of the field model shown in the fourth row of Figure~\ref{fig:long}.

This categorization regime appears to have the best hope of capturing the features of stable phonological distinctions in a linguistic context, and so we explore its behaviour further for two phonetic variables and multiple words in the next section.

 \section{Model and Simulation: Two Phonetic Variables} \label{sec:twodim}
 
To demonstrate the robustness of our models  we show that our qualitative results do not change dramatically  when exemplars are modelled as consisting of two phonetic variables rather than just one.
 The shift from one phonetic dimension to two  allows us to compare our results qualitatively  to the experimental data shown in Figure~\ref{fig:peterson_barney}, where the F1 and F2 of spoken exemplars are shown.

The models we presented in Subsections \ref{subsec:marmot1} and  \ref{subsec:marmot2} were for just one phonetic dimension, but they extend immediately to more than one dimension.
  For this section, we just interpret the variables  $y$ and $z$ as vectors of length two. The only proviso is that in \eqref{eqn:hyrax1} and \eqref{eqn:hyrax2} we interpret  $| y-z |$ as the Euclidean norm of $y-z$.  In the exemplar model, each exemplar still has a phonetic variable $y_i$ and a weight $w_i$, but now $y_i$ is a vector of length 2 rather than a scalar. In the field model, as before, we obtain equations for exemplar density fields $\rho_A(y,t), \rho_B(y,t), \ldots$ but now the dependent variable  $y$ is two-dimensional.

We show the results of two simulations, one exemplar model and one field model. Both simulations were done with the 5 words and the competition with discards regime for categorization. The parameters in the model are set as follows for both simulations: $\lambda=1$, $\mu=1$, $\sigma=1$, $\beta=0.1$, $p=1$, $k=10$, $y^*=(0,0)$.

\begin{figure}[h!]
\begin{center}
\includegraphics[width=6cm]{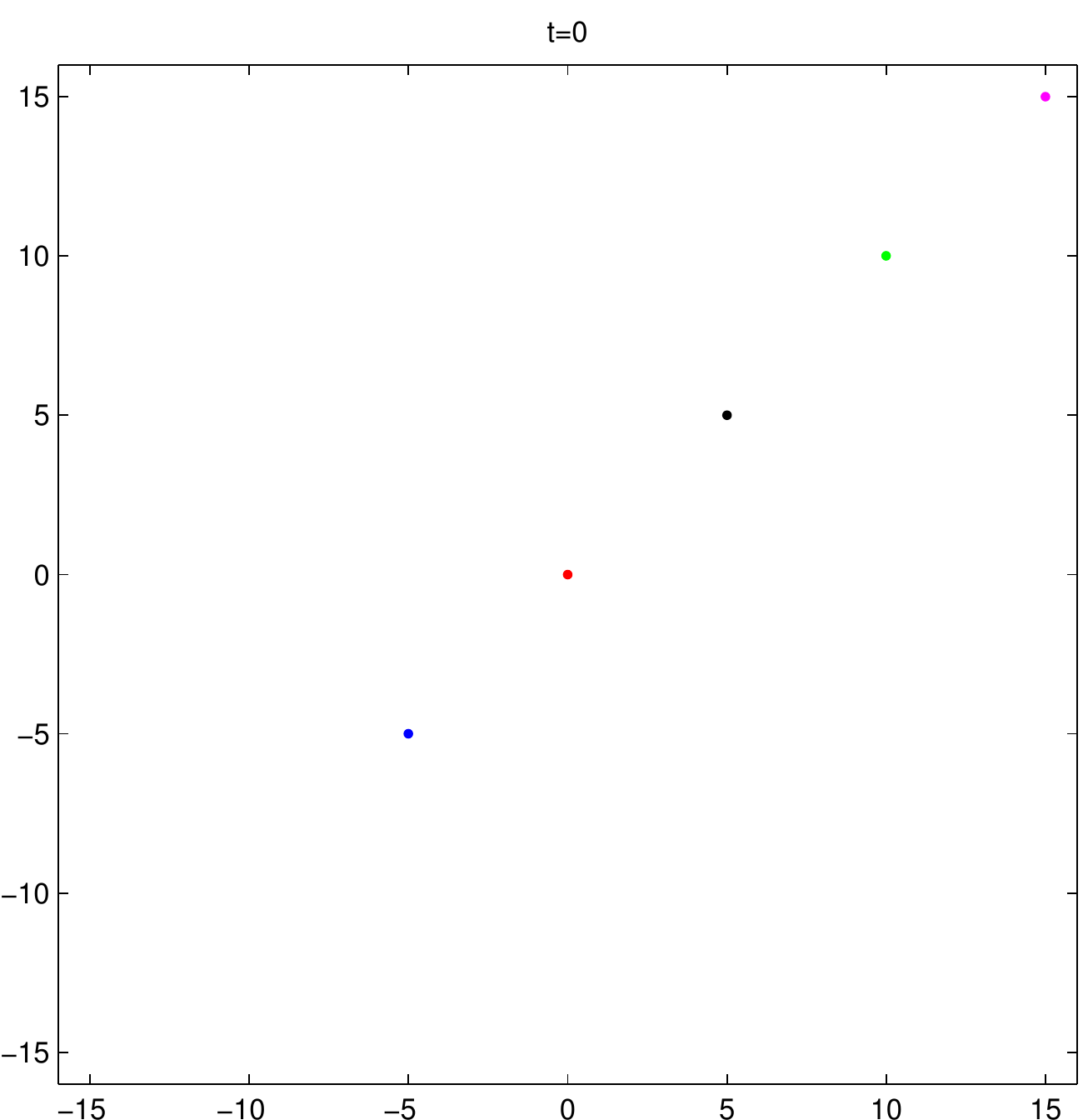}
\includegraphics[width=6cm]{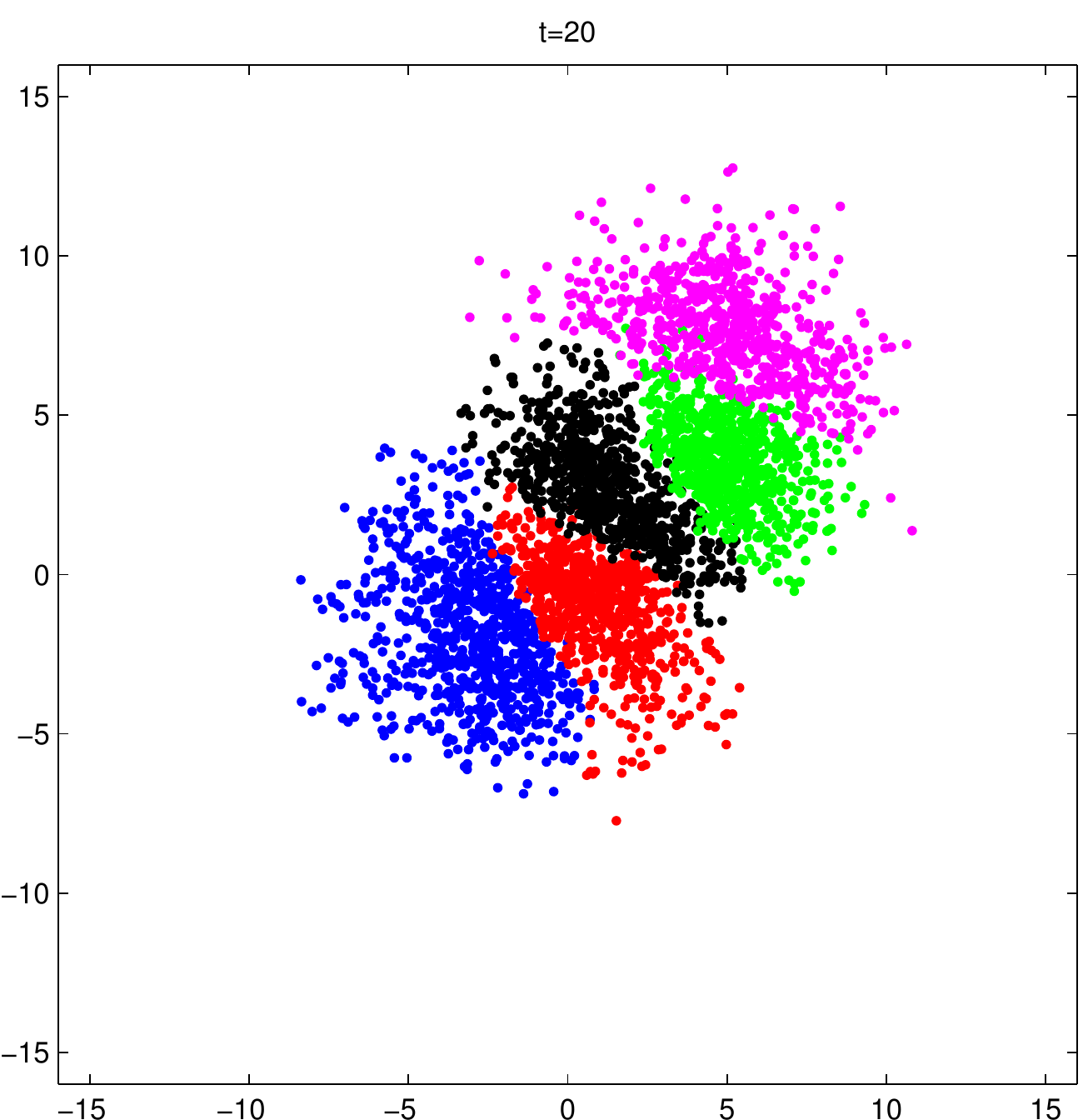}
\includegraphics[width=6cm]{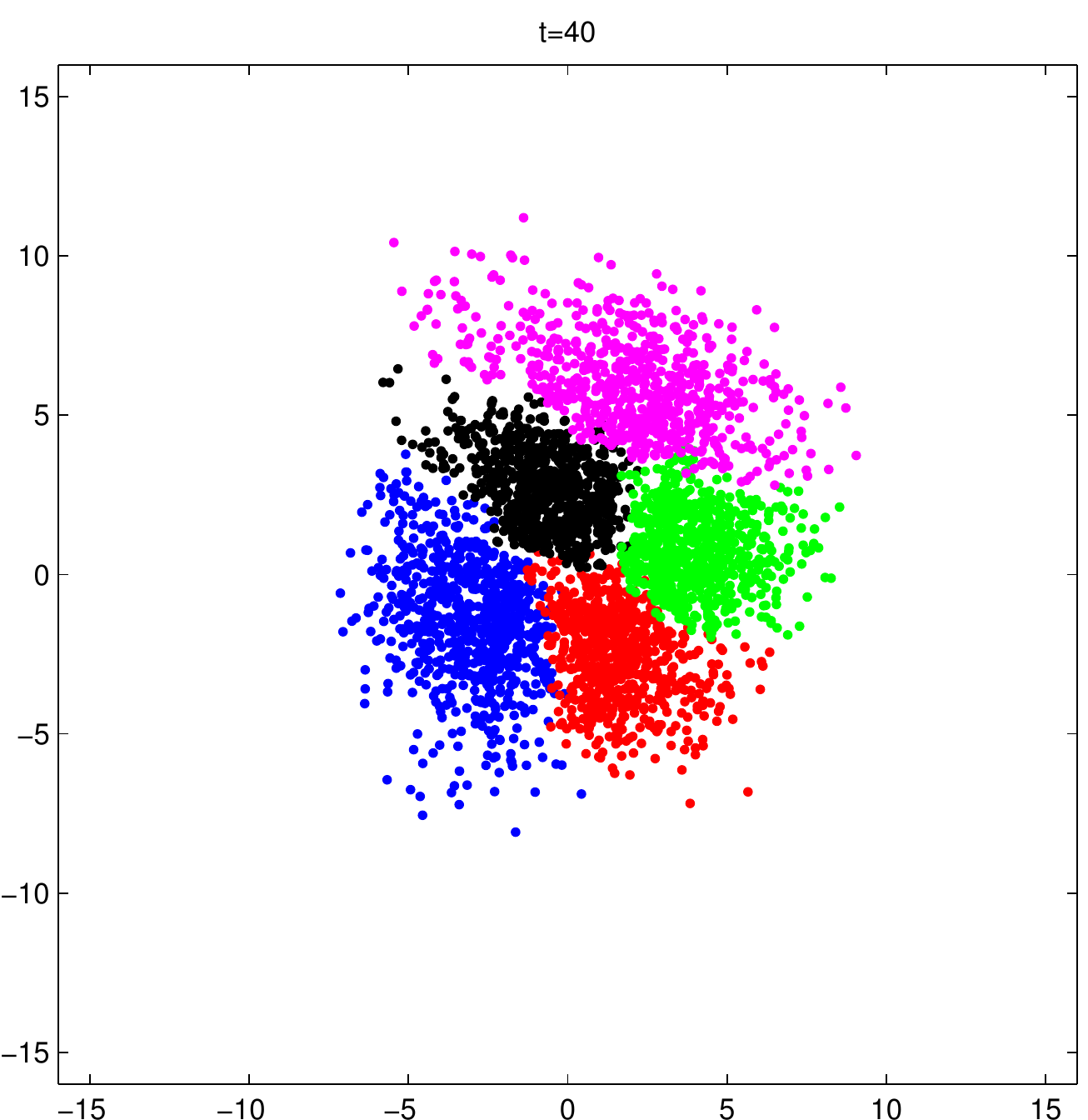}
\includegraphics[width=6cm]{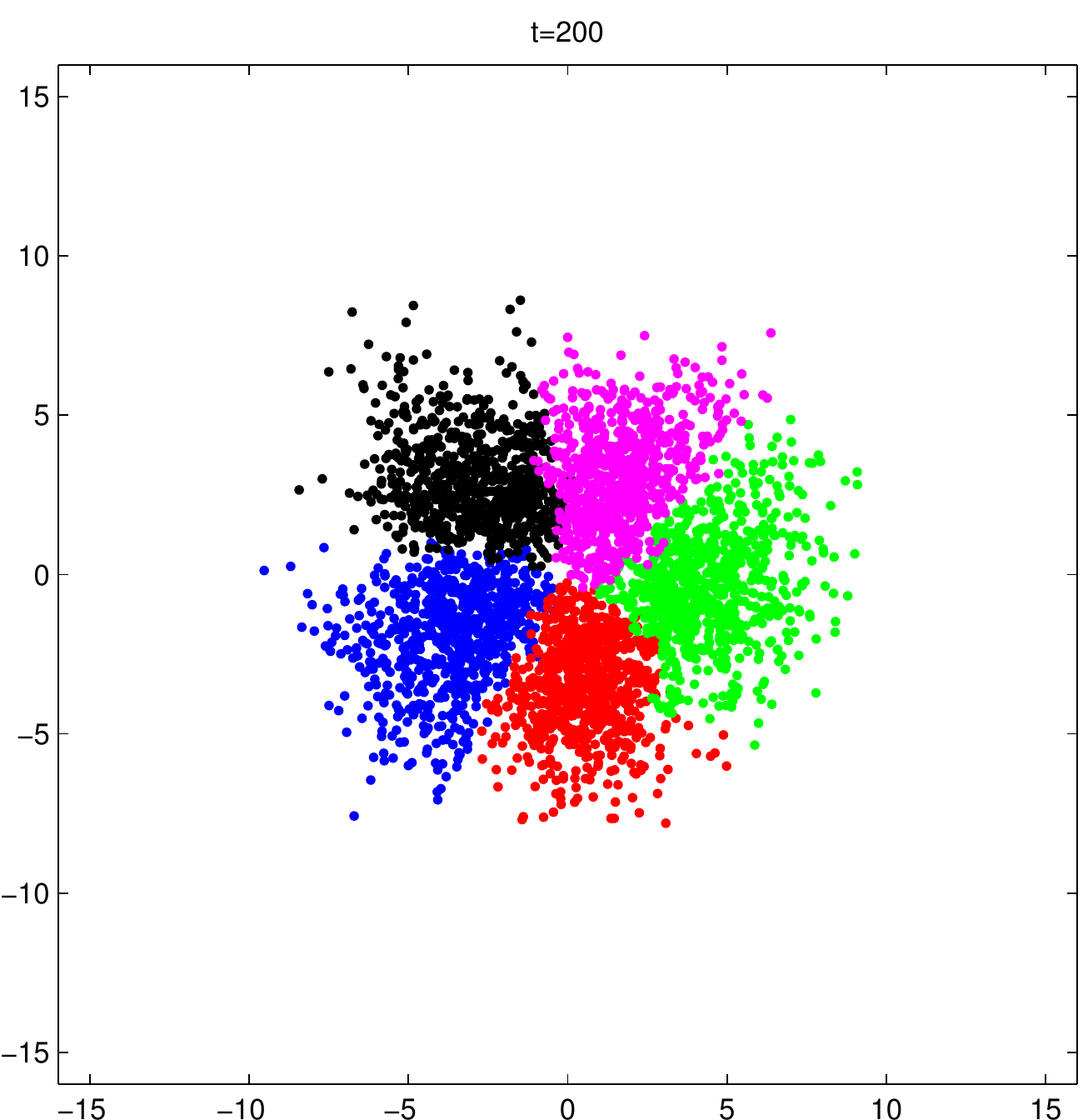}
\end{center}
\caption{ \label{fig:twodexemplar} The distribution of exemplars at four different times
in a simulation of the exemplar model in two dimensions with
6 words. The bottom right plot is representative of the
equilibrium distribution of the system.}
\end{figure}

In Figure~\ref{fig:twodexemplar} we show the results of the exemplar simulation. The system was initialized with 50 exemplars at the same location for each word, each of weight $10^{-3}$. The five sets of exemplars were located at the points $(-5,-5), (0,0), \ldots, (15,15)$.  Exemplars were deleted when their weight decreased below $10^{-4}$. In Figure~\ref{fig:twodexemplar} we plot the existing exemplars at four different times in the simulation.

\begin{figure}[h!]
\begin{center}
\includegraphics[width=6cm]{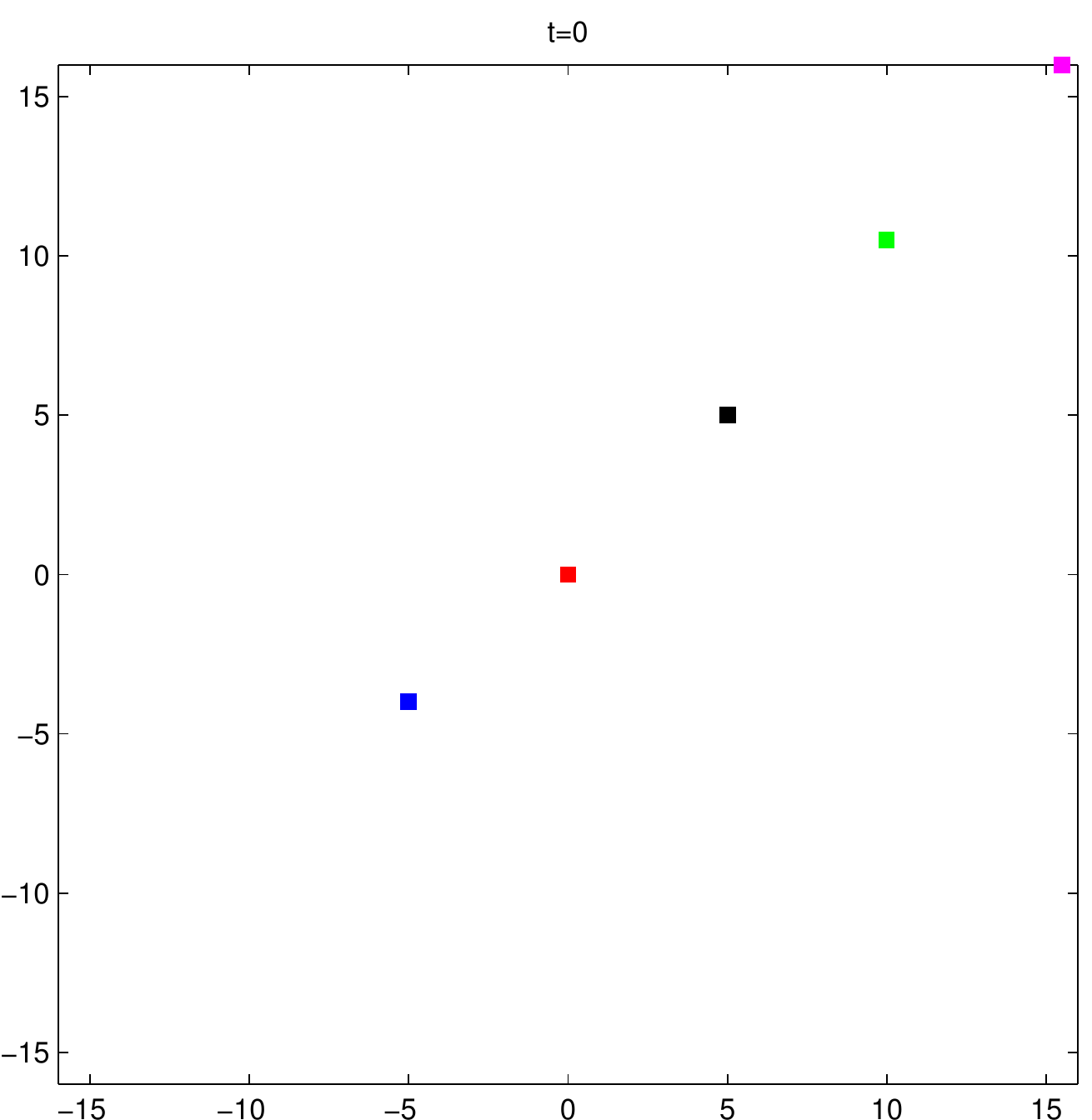}
\includegraphics[width=6cm]{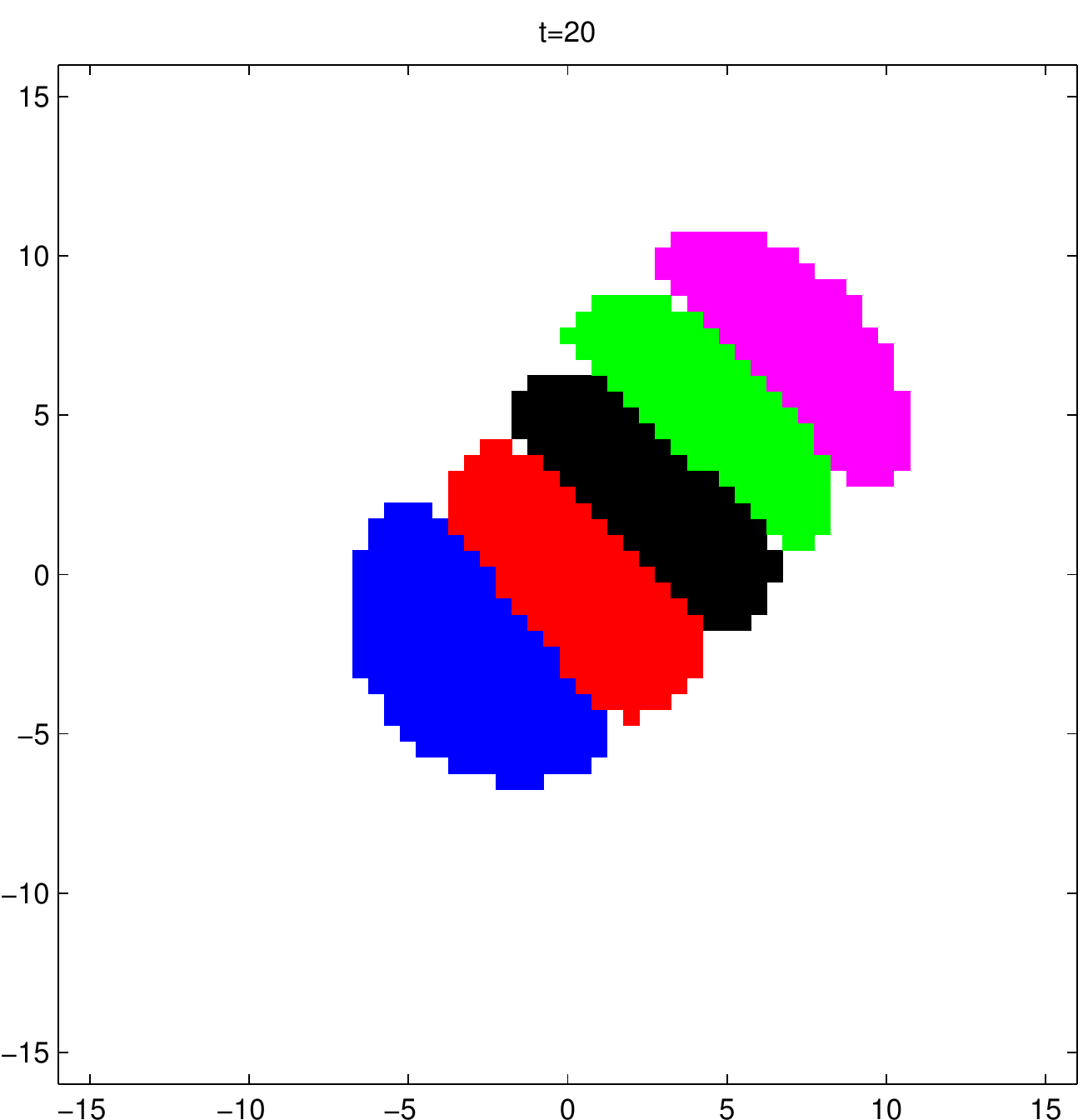}
\includegraphics[width=6cm]{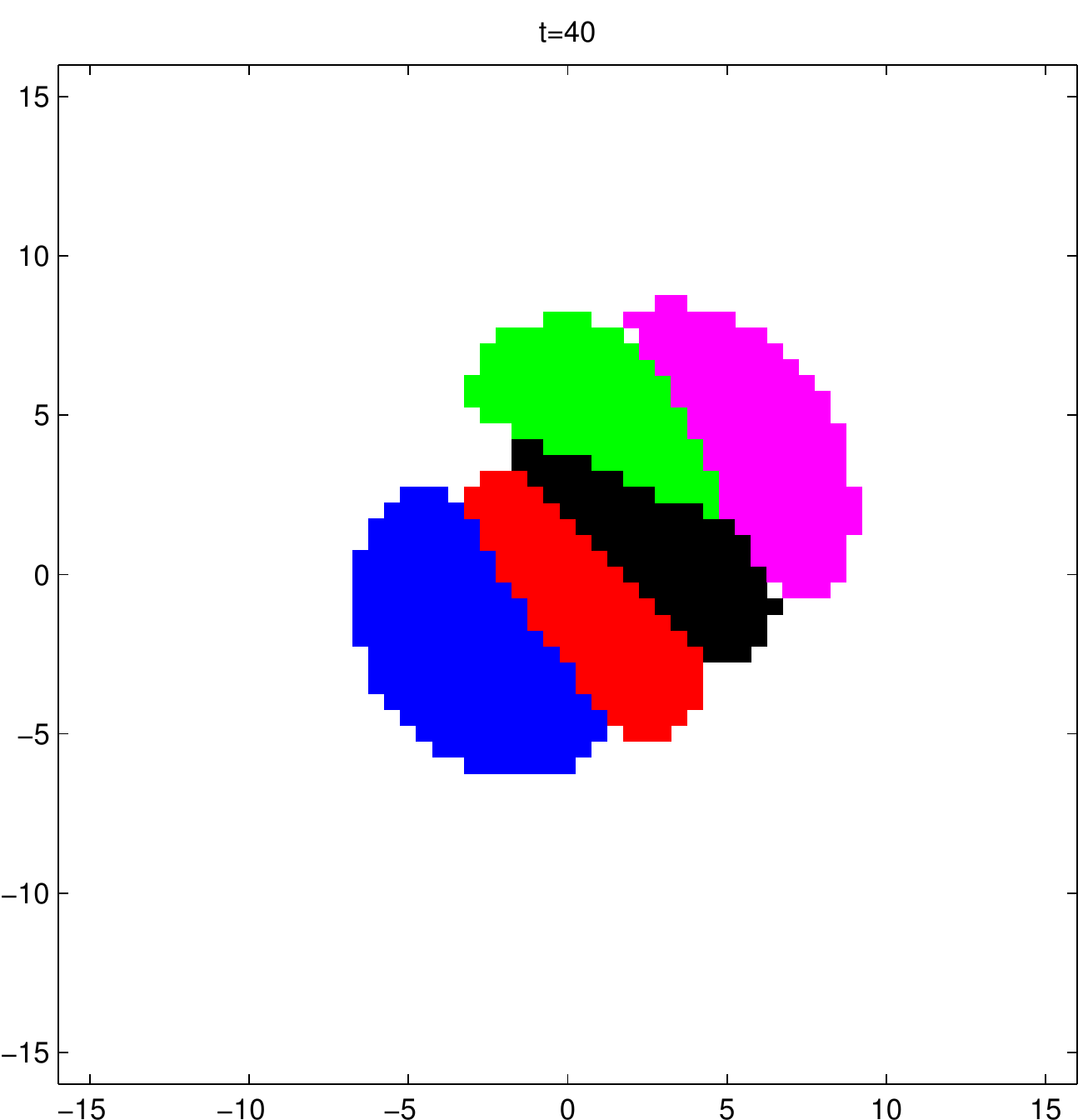}
\includegraphics[width=6cm]{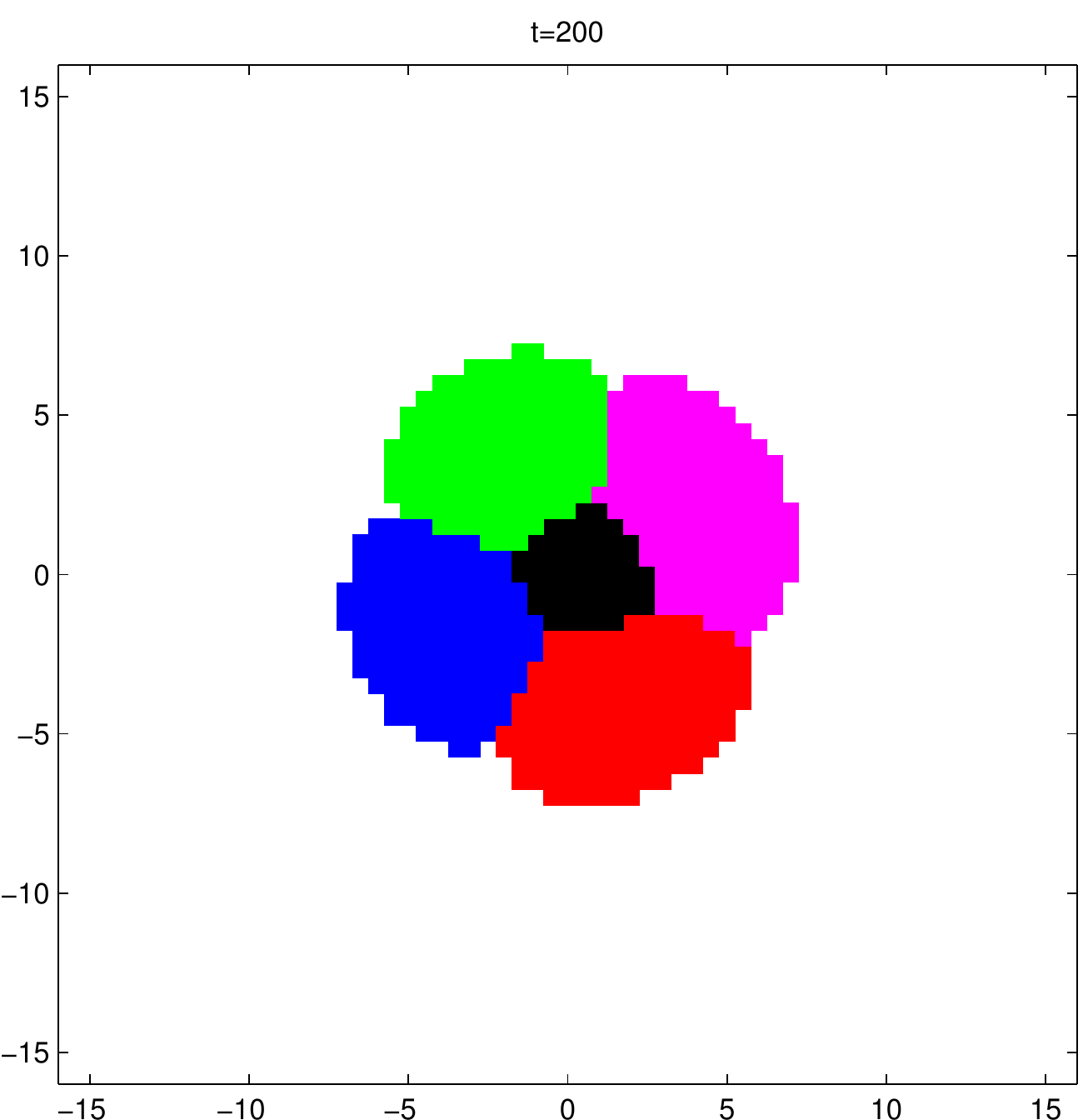}
\caption{ \label{fig:twodfield} Peaks in the exemplar density field at four different
times in a simulation of the field model in two dimensions
with 6 words. The bottom right plot is close to equilibrium.
}
\end{center}
\end{figure}

The corresponding simulation for the field model is shown in Figure~\ref{fig:twodfield}. 
Each field was initialized to zero except at a single point, giving the same initial total weight per word as in the exemplar simulation. These peaks in the initial conditions of the fields are located at the corresponding location in the exemplar simulation, except that we perturb them randomly by a small amount in order to break a symmetry that would otherwise persist throughout the simulation. In Figure~\ref{fig:twodfield}, at four different times, we plot the colour of the word with the maximum exemplar density $\rho$, except in regions where none of the $\rho$ have values greater than $0.001$.

We note that the  simulations of the two-dimensional exemplar and field models agree at a rough qualitative level, in terms of the size of the words' exemplar distributions and the time scale at which they settle to equilibrium. However, the configuration of the exemplars of different words about the origin in equilibrium differs, with the exemplar model producing something like a star, but the field model producing a flower-shape with one of the words at the centre.
This discrepancy is a least partly due to the randomness of the exemplar simulation and the randomness of the initial conditions of the field simulation. (In fact, other simulations of the field simulation with a different random seed yield a star pattern in equilibrium.) 

\section{Predictions of the Model: Functional Load} \label{sec:functional_load}

In our model we have imagined a language with only a few words whose pronunciation differ in only one sound.
As a consequence, when any two of the sounds merge, it leads to the the merger of precisely two words.  This is, of course, not representative of true languages.
Spoken languages make use of a fairly limited inventory sounds to create a vast collection of words. Recall that a \emph{minimal pair} consists of two words whose pronunciations differ in only one sound. We have tacitly assumed up till now that every pair of sounds distinguished exactly one minimal pair of words. In actual languages there are generally many minimal pairs for each pair of contrasting sounds.
  When the two sounds merge, the pronunciation of all these  minimal pairs  are merged. For example, in dialects of English where the {\sc cot--caught}  merger has occurred, not only do the words cot and caught merge, but also the pairs of words Don/Dawn and nod/gnawed, as well as many others \cite{labov2011}.
 
 What does our model predict for merger of sounds in real languages?
According to our model, whether two sounds merge or not would depend on how often there was  rejection of ambiguous exemplars for any of the minimal pairs distinguished by the two sounds. A very simple prediction then is that when a pair of sounds has no minimal pairs, and therefore no possible rejected exemplars, the two sounds will merge. Likewise, when a pair of sounds distinguishes many minimal pairs we expect that merger will not occur.  The number of minimal pairs that are contrasted by a pair of vowel sounds is known as the \emph{functional load} of the pair \cite{wedel_kaplan_jackson}. The \emph{functional load hypothesis} says that the propensity for a pair of sounds to merge is dependent on the function load of the pair, with greater functional load leading to a lower probability of merger (see, for example, \cite{hockett}). Like other models, including \cite{blevins_wedel}, our model predicts  that the functional load hypothesis will hold. This hypothesis is tested in \cite{wedel_kaplan_jackson}, where the authors perform a corpus study providing strong empirical support for this claim.

The above prediction is based on the assumption that whenever there is a minimal pair for a pair of sounds, there will be rejection of ambiguous exemplars of the words in the pair. But there is reason to believe that not all minimal pairs are equivalent in evoking exemplar rejection. In situations where context disambiguates which word in a minimal pair an exemplar is, we might expect that there is no rejection (or rarer rejection) of ambiguous exemplars. For example, if I hear the sentence ``You can sleep on the cXt.", where X is an ambiguous sound that could make the last word either ``cot" or ``caught", I can easily infer that ``cot" is the intended word, since ``the" cannot ordinarily be  followed by the word ``caught". More generally, because ``cot" is a noun and ``caught" is verb there are few contexts in which there is a genuine ambiguity between the words.
One might imagine that a minimal pair in which the words are both of the same part of speech (e.g. noun, verb, adjective) are more effective at preventing merger of sounds than minimal pairs consisting of pairs of words with different parts of speech.

Based on these kinds of considerations, many have predicted that rather than a  functional load that merely counts minimal pairs being the best predictor of whether two sounds merge, a functional load in which minimal pairs that are of the same part of speech are weighted more heavily than other minimal pairs would do a better job. This is in fact borne out by the corpus study \cite{wedel_kaplan_jackson_langandspeech}. Likewise, our model (and those of other researchers) predict that minimal pairs consisting of  more frequent pairs of words are more effective at predicting merger than those of rare words, and this is again supported by \cite{wedel_kaplan_jackson_langandspeech}.

\section{Conclusions and Discussion} \label{sec:conclusion}

We have provided a model of phonological contrast in which the choice of categorization regimes leads to either the merger of the pronunciation of two words  or to its maintenance of their difference. 
Our major finding is that the pure competition regime does not provide a good model of the preservation of phonological contrast.  When the selection parameter $p$ for this regime was set to  $1$, simulations showed that the pronunciations merged, albeit over a much longer time interval than for the no-competition regime. For $p>1$ there were distinct pronunciations, but they had a tendency to move in the opposite direction of the bias towards $y^*=0$ in the system.
  Our study provides support for the idea of some sort of exemplar rejection, as in our competition with discards regime
  and described by Labov  \cite[pp.\ 586--588]{labov} and Silverman \cite[Ch. 5]{silverman}.
In practice, which categorization regime best models the behaviour of hearers in different linguistic contexts is an empirical question that we do not address here.
We refer the reader to the discussion of these issues in \cite[Sec.\ 2.5]{wedel2012}

Another result  we found is that  including a bias towards the mean $\bar{y}$ of a category ($\alpha>0$, called entrenchment in \cite{pierrehumbert_exemplar}) was not necessary to preserve phonological contrast. A combination of the pull towards $y^*$ (the phonetically preferred value of $y$)  together with the interaction between exemplars of the two words was enough to prevent the exemplar distribution from spreading out indefinitely. Of course, this in no way precludes entrenchment from being a real phenomenon in actual language use. 

Finally we note that,  comparing our simulated results to the data in Figure~\ref{fig:peterson_barney},  a striking difference is that the distribution of exemplars  in the experimental data do not fill a roughly circular space, as it does in our results. In our model we have assumed that phonetic space is homogenous except for a bias towards the origin.  Articulatory and acoustic data indicate that there are plenty of inhomogeneities in phonetic space. A better model of the range of allowable vowels, as considered by \cite{boer} and \cite{jaeger} is that of a triangle or rectangle, whereas our model imposes no restrictions on utterable sounds but the bias of the speaker towards the centre of the phonetic space causes the used domain to be roughly circular. We could straightforwardly modify our model to take some of these factors into account but we have not done so here.

\section*{Acknowledgments}
The authors thank Andy Wedel, J. F. Williams, John Alderete, Ben Goodman, Dan Silverman, Nilima Nigam, Kathleen Hall, and Molly Babel for comments on an earlier draft.

\def\cprime{$'$} \def\cprime{$'$}

\bibliographystyle{siam}

\end{document}